\documentclass{article}

\usepackage[preprint]{neurips_2026}

\usepackage[utf8]{inputenc} 
\usepackage[T1]{fontenc}    
\usepackage{hyperref}       
\usepackage{url}            
\usepackage{booktabs}       
\usepackage{amsfonts}       
\usepackage{nicefrac}       
\usepackage{microtype}      
\usepackage{xcolor}         

\usepackage{microtype,graphicx,subcaption,enumitem,hyperref,amsmath,amssymb,mathtools,amsthm,pgfplots,pgfplotstable,tikz,textcomp,pifont,adjustbox,multirow,wrapfig,makecell}
\pgfplotsset{compat=1.18}
\usepgfplotslibrary{groupplots}
\usepackage{siunitx}
\usepackage{pgfplotstable}
\usepgfplotslibrary{fillbetween}
\usepackage{siunitx}
\usepgfplotslibrary{statistics}
\usepackage{caption}
\title{Geometry-Centered 3D Latent World Models for Growing Surfaces}

\author{%
  Xiaoyi Liu \\
  Department of Computer Science\\
  Brown University\\
  \texttt{xiaoyi\_liu@brown.edu} \\
  \And
  Hao Tang$^\dagger$ \\
  School of Computer Science \\
  Peking University \\
  \texttt{haotang@pku.edu.cn} \\
}

\begin{document}

\maketitle
\begingroup
\renewcommand{\thefootnote}{\fnsymbol{footnote}}
\footnotetext[2]{Corresponding author.}
\endgroup

\begin{abstract}
Many physical systems do not merely move or deform; they grow, adding material and changing the geometry that a world model must represent.
Existing world models are typically optimized for pixel prediction, reward prediction, or fixed-support physical dynamics, leaving open how to model systems whose underlying physical support expands over time and whose future morphology depends on hidden material response.
We introduce FOLIAGE, a geometry-centered latent world model for growing surfaces.
Within a fixed state budget, FOLIAGE represents mature regions as a compact scaffold while allocating higher-resolution state to regions predicted to drive near-future growth.
This focuses representation and computation where new material and geometric change occur while retaining compact global context.
FOLIAGE further separates observation, action, and privileged physics: heterogeneous RGB, point-cloud, and mesh observations are fused into a deployable geometric state; material controls condition the latent dynamics; and hidden physical energies guide training but are not required at deployment.
To evaluate this setting, we introduce SURF-GARDEN and SURF-BENCH, providing controlled counterfactual branches, dense cross-modal correspondences, hidden physical signals, and stress tests for growing-geometry state learning.
FOLIAGE reduces inverse-material error by $\approx40\%$ and 5-step mesh forecasting Chamfer error by $\approx30\%$ relative to strong baselines, while improving cross-modal retrieval by +14 mAP points.
Stress tests show graceful degradation under sensor loss and correspondence corruption.
On temporal 3D plant scans, FOLIAGE also improves passive future-geometry forecasting, while transfer experiments show that the learned geometry-centered state remains useful beyond the simulator.
\end{abstract}

\section{Introduction}
Many physical systems take shape by growing. Leaves curl, tissues fold, and active or printed sheets change form as local material response varies across a surface. 
In such systems, geometry is not merely displaced; the surface itself expands, new degrees of freedom appear, and local growth can propagate into global morphology. 
Modeling this process matters because \textit{growth is a mechanism for creating form and function}: it determines how biological structures develop, how soft materials bend and fold, and how programmable surfaces can be designed to reach desired shapes in fabrication. \cite{liang2011growth, Coen2023PlantMorphogenesis, Efrati2017_ShapeShifting, Gladman2016Biomimetic4D, Wang20244DCeramics, vanManen20214DMetamaterials, Ambrosi2019GrowthRemodelling}. 
It also poses a fundamental machine-learning question: how should a learned world model represent a physical state whose geometry grows over time? 
We study this question in thin growing surfaces, where local changes in surface material can be observed as large-scale changes in shape.
This setting is simple enough to define geometrically, yet rich enough to require tracking where new surface appears, how older and newer regions interact, and how hidden material response shapes future geometry.

Related approaches partially address this problem, but begin from different state assumptions. Video and latent world models learn how observations evolve over time, so they can capture visible evidence of growth \cite{Hafner2019_PlaNet,Hafner2023_DreamerV3}. For growing surfaces, however, visible change is only a partial view of physical change: internal stress that shapes the surface is hidden, and an active region may curl, self-occlude, or leave the current view while continuing to affect future geometry. Physics-based simulators and learned physical dynamics model the evolution of physical states represented by meshes, particles, or fields \cite{hu2021diffpd,li2022diffcloth}. These approaches are powerful when the current physical state and discretization are given. Growing surfaces add a complementary state-estimation problem: the current geometry may be only partially observed, while future material support, birth locations, and connectivity are not known in advance. Static 3D representation and correspondence methods track geometric identity, but do not model how growth changes under material interventions \cite{melzi2019zoomout,Sharp2022_DiffusionNet}. These perspectives suggest the state abstraction needed for growing geometry: \textit{a geometric representation of the evolving surface that can be inferred from heterogeneous partial observations, updated under material controls, and shaped during training by hidden physical signals when available.}

\begin{figure*}[t]
  \centering

  \includegraphics[width=0.98\textwidth,height=5cm,keepaspectratio]{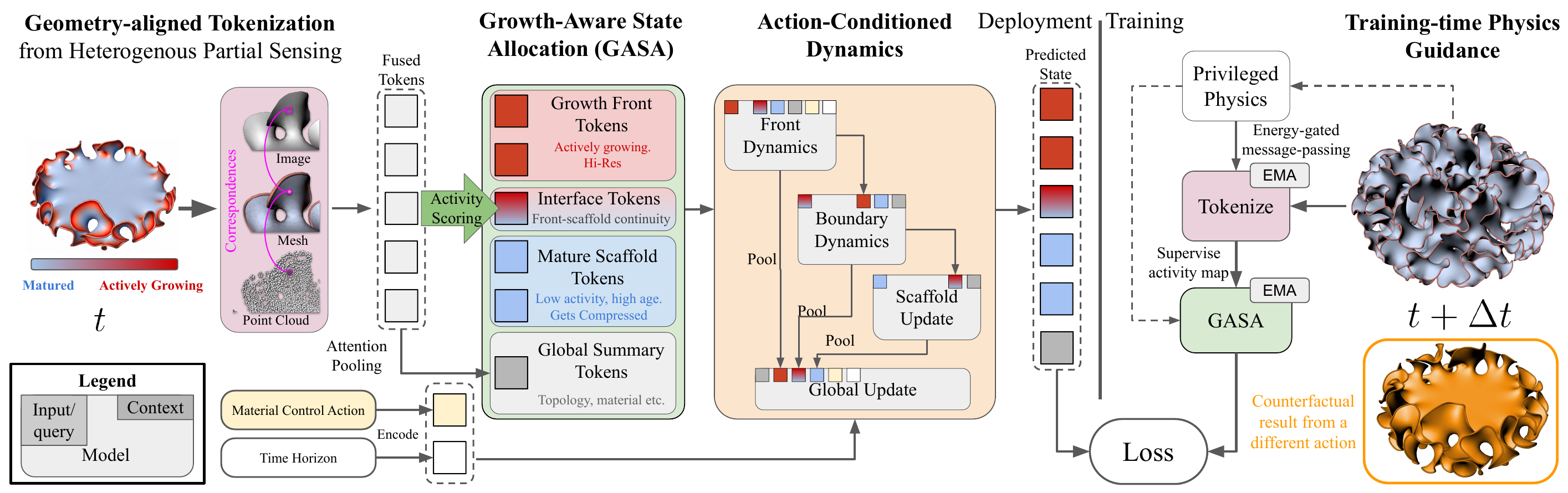}

  \caption{FOLIAGE models growing surfaces by building a geometric state from partial sensors, allocating a fixed state budget according to estimated growth activity, and rolling the state forward under material-control actions. Privileged physical signals guide training only (not required at deployment).}
  \label{fig:model-arch}
  \vspace{-0.6cm}
\end{figure*}

We introduce FOLIAGE, a geometry-centered latent world model for growing surfaces (Fig.~\ref{fig:model-arch}). Its central algorithmic idea is \textit{growth-adaptive state allocation} coupled with structure-matched dynamics. The physical surface represented by the model can gain material and geometric degrees of freedom over time, but representing this growth does not require the latent state to grow with the mesh. Instead, FOLIAGE uses a fixed state budget and assigns that capacity according to where future geometric change is expected. Mature regions are represented as a compact scaffold that preserves global shape, topology, boundary conditions, and long-range constraints. Young or high-activity regions receive denser growth-front tokens, while interface tokens preserve continuity between the active front and the mature surface. The dynamics model then updates these roles differently, spending its most expressive computation on future-active regions while retaining the surrounding scaffold as context. This gives a practical inductive bias for growing surfaces: as the underlying geometry expands, state capacity should follow the parts of that geometry that matter most for its future evolution.

The next challenge is to preserve the state representation's counterfactual meaning: from the same observed surface, the model should predict different futures under different material controls. FOLIAGE therefore keeps three roles separate. Observations from RGB images, point clouds, and meshes infer the current growing-surface state. Material controls enter only through the latent dynamics, so they intervene on the future rather than changing the meaning of the current state. Hidden physical quantities, such as per-vertex energies that summarize internal stress, guide representation learning during training but are excluded from the deployed path. This role separation lets FOLIAGE use privileged physics to learn better states while preserving a practical observation-based model for action-conditioned geometric prediction. In particular, one inferred current state can be queried under multiple material controls, so the action selects a possible future rather than changing the representation of the observed present.

To evaluate growing-geometry world models, visual prediction alone is insufficient. A useful state should preserve global geometry, track identity through growth, align heterogeneous sensors, infer hidden material causes, predict counterfactual futures, and remain usable under partial or noisy observations. We therefore introduce SURF-GARDEN, a controlled data platform for generating growing surface sequences, and SURF-BENCH, an evaluation suite organized around these requirements. SURF-GARDEN provides counterfactual branches, dense temporal and cross-modal correspondences, and hidden physical signals that are difficult to measure at scale in real growth data. The task suite probes whether the latent behaves like a physical state of a growing object: topology and dense correspondence test geometry understanding; image--point retrieval and sensor ablations test multimodal grounding; inverse material estimation and growth-stage recognition test hidden physical fidelity; and mesh forecasting and long-horizon rollout test action-conditioned dynamics. Across these tasks and stress tests, FOLIAGE outperforms task-specific baselines while degrading gracefully under sensor and correspondence corruption. Real plant experiments additionally execute the learned transition model for passive future-geometry prediction, while plant and dynamic-human transfer experiments test whether the learned representation remains useful beyond SURF-GARDEN.

In summary, our contributions are:
\begin{itemize}
    \item \textbf{Growing geometry as a world-modeling problem.}
    We formulate growing surfaces as a regime where the underlying physical support changes over time, current geometry must be inferred from partial observations, material controls alter future morphology, and future support is not known in advance.

    \item \textbf{Growth-adaptive geometric state and structured dynamics.}
    We introduce FOLIAGE's central algorithmic mechanism: a fixed state budget is organized into mature scaffold, active growth-front, and interface tokens, with role-specific dynamics that concentrate representation and computation on predicted future-active regions.

    \item \textbf{Counterfactual interface with train-only physics.}
    FOLIAGE forms the current state from observations, introduces material interventions in the transition, and confines privileged physical signals to training. This allows one observed state to support multiple action-conditioned futures without requiring solver quantities at deployment.

    \item \textbf{Controlled benchmark for growing-geometry state learning.}
    We introduce SURF-GARDEN and SURF-BENCH, providing counterfactual branches, dense correspondences, hidden physical signals, and stress tests for evaluating geometry, identity, multimodal grounding, physical fidelity, and predictive dynamics.

    \item \textbf{Empirical evidence across geometry, dynamics, robustness, and real-world forecasting.}
    FOLIAGE improves geometric understanding, multimodal grounding, hidden-material inference, and forecasting; matched-budget controls isolate the benefit of growth-aware allocation and structured dynamics; and temporal plant scans demonstrate passive future-geometry prediction beyond the simulator.
\end{itemize}

\section{Related Works}

\textbf{Video and latent world models.}
Latent world models learn compact states for prediction, planning, and control from high-dimensional observations \citep{Ha2018_WorldModels,Hafner2019_PlaNet,Hafner2023_DreamerV3}. Most work focuses on image-centric settings, where the state is trained via reconstruction or policy objectives. In parallel, neural scene representations and 3D perception emphasize explicit spatial structure for view synthesis, reconstruction, and embodied reasoning \citep{Mildenhall2020NeRF,Kerbl2023_3DGS,Jaegle2021_Perceiver}. FOLIAGE follows this geometry-centered direction but targets a distinct regime: geometry is not just a representation, but the evolving physical state. Growing surfaces require tracking material identity, support, and future morphology under control, making geometry itself the object of prediction.

\textbf{Learned physical dynamics and growing support.}
Differentiable simulators and neural physical models enable inverse problems, control, and forecasting when states and assumptions are specified \citep{Hu2020DiffTaichi,Hu2019ChainQueen,hu2021diffpd,li2022diffcloth}. Graph-based simulators and operator-learning methods further model dynamics over particles, meshes, and fields with strong generalization \citep{SanchezGonzalez2020_GNS,Pfaff2021_MeshGraphNets,Li2021_PINO}. However, growing surfaces introduce changing state support and partially observed internal physics. FOLIAGE addresses this by learning latent dynamics over geometry, using simulator signals only during training. This preserves physical structure without requiring solver access at deployment.

\textbf{Multimodal geometry, correspondence, and physical identity.}
Multimodal 3D learning aligns images, point clouds, and meshes for retrieval and recognition \citep{zhang2021crosspoint,zhu2022pointclip,li2023ulip2}, while correspondence and functional-map methods match surfaces across shapes \citep{ovsjanikov2012functional,melzi2019zoomout,Donati2020DeepGeomFMaps,Sharp2022_DiffusionNet}. In growing geometry, correspondence extends beyond static matching to track physical identity as surfaces evolve. FOLIAGE uses correspondence-aware fusion to build a shared geometric state from partial observations, ensuring it supports both static tasks and future prediction under growth.

\textbf{Simulation, counterfactual data, and privileged training signals.}
Simulation provides scalable supervision, controlled interventions, and privileged variables \citep{tobin2017domain,james2019sim,makoviychuk2021isaac,szot2021habitat}. This is especially valuable for growing geometry, where counterfactual futures, dense correspondences, and internal energies are difficult to observe. Related ideas in privileged learning and distillation leverage additional training signals to improve deployment representations \citep{vapnik2015priv,hinton2015distilling,tarvainen2017mean}. FOLIAGE follows this paradigm: per-vertex energies guide training, while deployment relies only on observable sensors. SURF-GARDEN and SURF-BENCH provide controlled evaluation with counterfactual growth and robustness tests.

\textbf{Growing surfaces and morphology.}
Growth-driven shape formation has been widely studied in mechanics, morphogenesis, and active materials \citep{efrati2009elastic,liang2011growth,Gladman2016Biomimetic4D,vanrees2017growth,Ambrosi2019GrowthRemodelling}. These works show how local growth and material response produce complex global morphology. FOLIAGE brings this regime into learned world modeling, focusing on inferring compact states from partial observations and predicting their evolution under control. More broadly, when systems gain degrees of freedom over time, effective world models benefit from growth-adaptive state allocation that separates stable structure from regions of active change.

\section{FOLIAGE: A Growth-Adaptive World Model for Growing Surfaces}

\subsection{Problem formulation and interfaces}

FOLIAGE uses a world-model interface specialized to growing surface geometry. At time $t$, the model receives any \textit{subset} of RGB images, point clouds, and surface meshes, $x_t \subseteq \{I_t,P_t,M_t\},$
and infers a current geometric state $S_t$. Given a material action $a_t \in \mathbb{R}^3$ and prediction horizon $\Delta t$, its transition model predicts $\hat{S}_{t+\Delta t}=P_\theta(S_t,a_t,\Delta t).$
When no intervention is supplied, a learned null-action token supports passive prediction. The growing-surface setting changes what this interface must represent: the future surface may contain material and geometric degrees of freedom that do not exist at time $t$, and its future support, birth locations, and connectivity are not provided to the model.

Importantly, changing physical support does not require the number of latent tokens to increase with the surface. FOLIAGE instead uses fixed scaffold, front, and interface budgets. At each encoded observation, growth-adaptive state allocation determines which parts of the observed surface receive detailed state. The transition then predicts a future structured state supervised by a target state formed from the future, expanded surface. Thus, the amount of represented physical material may increase substantially from $t$ to $t+\Delta t$, while the latent representation remains computationally controlled and changes what geometric information its capacity represents.

\textbf{Abstraction.}
FOLIAGE represents the current surface by the structured state
$S_t =
\{
s_t^{\mathrm{global}},
S_t^{\mathrm{mat}},
S_t^{\mathrm{front}},
S_t^{\mathrm{bdry}}
\}.$
$s_t^{\mathrm{global}}$ summarizes the whole surface, $S_t^{\mathrm{mat}}$ stores compact tokens for mature scaffold geometry, $S_t^{\mathrm{front}}$ stores higher-resolution tokens for regions predicted to drive near-future change, and $S_t^{\mathrm{bdry}}$ stores interface tokens connecting the active front to the scaffold. This decomposition is the central interface of FOLIAGE: growing geometry requires stable global context together with expressive local state where new material and deformation are likely to appear.

The method is organized around four subproblems. First, heterogeneous observations are converted into geometry-aligned tokens (Sec.~\ref{sec:method-p1}). Second, those tokens are allocated into mature scaffold, active front, and interface state (Sec.~\ref{sec:method-p2}). Third, the structured state is updated under material controls while keeping current-state inference observational (Sec.~\ref{sec:method-p3}). Fourth, hidden physical quantities improve training without becoming test-time requirements (Sec.~\ref{sec:method-p4}).

\subsection{Build a geometric state from heterogeneous partial observations} \label{sec:method-p1}

A growing-geometry world model needs a state whose coordinates are geometric rather than purely visual. Since observations may arrive as RGB images, point clouds, meshes, or subsets thereof, FOLIAGE first maps each available modality into a shared token space $U_t = T_I \cup T_P \cup T_M.$
Missing modalities contribute empty token sets, so downstream modules operate on the same representation regardless of which sensors are present. Image tokens provide appearance cues, point tokens provide metric anchors, and mesh tokens provide surface structure.

The part of tokenization most specific to growing geometry is the mesh pathway. Mesh vertices are not all equivalent: some correspond to older material that mainly anchors the global scaffold, while others were recently added and are more likely to indicate active growth. Each mesh token therefore carries geometric features and a birth-time tag $\tau(v)$, from which we derive an age feature
$\alpha_t(v) = t - \tau(v).$
Image and point tokens inherit age information through nearby mesh correspondences when available. This lets the model reason about the age structure of the surface even when the current observation is an image or point cloud.

FOLIAGE fuses modalities through a sparse correspondence graph rather than dense all-to-all attention. The graph contains image-to-mesh edges from projected visible surface elements, point-to-mesh edges from local geometric proximity, and mesh adjacency edges from surface connectivity. Type-aware attention over this graph lets image appearance refine geometric tokens and point measurements supply metric evidence, while keeping all modalities tied to the same physical surface. Because practical sensing can be partial or misaligned, training uses token masking, paired masking along correspondence edges, and modality dropout. Thus, correspondences are used as a useful geometric interface, not as a deployment assumption that every observation is complete or perfectly registered. We test this in Sec.~\ref{sec:results-deployment}

\subsection{Allocate state capacity according to growth} \label{sec:method-p2}

Growing surfaces are not uniformly dynamic. Mature regions often preserve global shape, topology, and boundary context, while material addition and control-sensitive deformation concentrate near active growth fronts. A single pooled latent risks hiding these differences: it may underrepresent the active regions where the future changes, or spend unnecessary capacity on mature regions whose near-term dynamics are small. FOLIAGE therefore performs \textbf{Growth-adaptive State Allocation (GSA)}: it compresses mature geometry into scaffold tokens, retains high-resolution tokens for active growth fronts, and adds interface tokens that keep local growth attached to global structure. Here, ``higher resolution'' refers to allocating a denser share of the fixed latent-state budget, together with a more expressive dynamics update, to predicted future-active regions. It does not refer to increasing the resolution of the output mesh itself.

Given fused tokens $U_t$, FOLIAGE first predicts an observable activity score
$g_i = \sigma\big(f_\psi(u_i)\big)$
for each token. The score estimates where near-future change is likely. It is computed from observable token features, including geometry, age, modality, and correspondence context. During training, hidden physical quantities and future local change can supervise this score, but they are not inputs to the deployed selector. This preserves the main interface: FOLIAGE can learn from privileged physics while still forming its state from observable sensors alone.

The activity scores define three complementary token sets for $S_t$. Low-activity or older regions are coarsened into \emph{mature scaffold tokens} $S_t^{\mathrm{mat}}$, which preserve global geometry, topology, and long-range constraints. High-activity regions become \emph{growth-front tokens} $S_t^{\mathrm{front}}$, which receive the highest-resolution representation because they are where new degrees of freedom are most likely to appear. Tokens near the transition between front and scaffold form \emph{interface tokens} $S_t^{\mathrm{bdry}}$, which preserve attachment between local growth and the surrounding surface. Finally, attention pooling over all fused tokens produces $s_t^{\mathrm{global}}$, a compact summary of overall shape and growth stage.

This decomposition is the main algorithmic mechanism of FOLIAGE. As the physical surface gains new support, the model must decide where a fixed amount of state capacity is most useful. GSA makes this decision from observable evidence: active regions receive detailed state, mature structure remains compact, and interface tokens preserve their attachment. The same structure then determines how dynamics are applied, with the strongest updates at the front and progressively lighter updates toward the mature scaffold. In Sec.~\ref{sec:results-ablation}, we compare against random and incomplete decompositions as well as same-budget uniform sampling, learned latent queries, generic learned pruning, and growth-aware allocation with homogeneous dynamics. These controls separate the effect of growth-specific allocation from additional capacity or generic adaptivity.

\subsection{Predict future geometry under material control actions} \label{sec:method-p3}

A useful model for growing surfaces should support counterfactual prediction: the same observed state can evolve differently under different material controls. FOLIAGE therefore separates \textbf{observation} from \textbf{action}. The perception and allocation modules form $S_t$ using only observable inputs; material controls enter only in the dynamics model:
$\hat{S}_{t+\Delta t} = P_\theta(S_t, a_t, \Delta t).$
This separation preserves the state's meaning: $S_t$ describes the current growing surface, and $a_t$ selects a possible future.

The dynamics model respects the structure introduced by GSA. Active growth-front tokens receive the most expressive action-conditioned update, because they represent regions where new material and near-future deformation are concentrated. Interface tokens propagate front changes into the surrounding surface, while mature scaffold tokens receive a lightweight residual update that preserves global geometry and long-range context. This gives the model a simple inductive bias: actions should influence the parts of the surface most likely to change, but those changes should remain attached to the mature scaffold. Material action $a_t$ is encoded as an action token, and the horizon $\Delta t$ a time token. When material control actions are unavailable, a learned null-action token supports passive prediction. Multi-horizon training exposes the model to short and longer prediction intervals, allowing the same transition model to support direct counterfactual prediction and recurrent rollouts.

\subsection{Use hidden physics as train-time guidance} \label{sec:method-p4}

The future of a growing surface depends on internal stress and material response, but these quantities are generally not available at deployment. FOLIAGE therefore separates \textbf{privileged physics} from both observation and action: hidden physical signals guide representation learning during training, while the deployed model uses observable inputs alone.

We maintain two encoders. The \textbf{context encoder} consumes only observable inputs (Sec.~\ref{sec:method-p1}) and produces the deployable state $S_t$. The \textbf{target encoder} is used only during training; it receives the same observations plus simulator-provided per-vertex physical energies and produces the target future state used to supervise prediction. The target encoder is updated as an exponential moving average of the context encoder, and prediction gradients do not flow through privileged physical inputs.

Physical energies are used in two ways. First, in the target encoder, they gate early mesh message passing so that high-stress regions contribute more strongly to the target representation. Second, they supervise the activity map used by GSA, together with future local change, teaching the observable context branch where growth-driven change is likely to occur. These signals improve the learned state without changing the test-time interface: at inference, FOLIAGE receives only $x_t$, forms $S_t$, and predicts future states under material actions.

We use several safeguards to keep the training/deployment separation explicit. Energies are detached constants, gates are bounded, physics-guided message passing is applied only in the target branch, and the target branch is EMA-coupled to the observable context branch. Thus, train-only physics shapes the representation, but the deployed model remains solver-free.

\subsection{Training and Deployment}

FOLIAGE is trained to predict the future structured state produced by the target encoder. The objective contains latent prediction, activity supervision, energy prediction, and representation regularization:
$\mathcal{L}
=
\mathcal{L}_{\mathrm{pred}}
+
\lambda_{\mathrm{act}} \mathcal{L}_{\mathrm{act}}
+
\lambda_{\mathrm{energy}} \mathcal{L}_{\mathrm{energy}}
+
\lambda_{\mathrm{vc}} \mathcal{L}_{\mathrm{vc}}.$
$\mathcal{L}_{\mathrm{pred}}$ matches the predicted future state to the target state, $\mathcal{L}_{\mathrm{act}}$ teaches the activity selector where future change occurs, $\mathcal{L}_{\mathrm{energy}}$ drives observable features to correlate with hidden stress, and $\mathcal{L}_{\mathrm{vc}}$ prevents representation collapse.
The inference path is simpler: observable inputs are tokenized, GSA forms the structured state $S_t$, and the dynamics model predicts $\hat{S}_{t+\Delta t}$ under material action $a_t$ and horizon $\Delta t$. No target encoder, physical energy, or simulator state is used. Downstream task heads read from the same structured state. 

For explicit geometry forecasting, a variable-cardinality readout converts $\hat{S}_{t+\Delta t}$ back to surface geometry. Queries anchored to the current surface predict updates to existing geometry, while learned growth queries predict candidate new vertices and validity scores; valid additions are reconstructed locally. Future vertex count, birth locations, and connectivity are therefore not provided as inputs. This readout tests whether the predicted state contains the information needed to reconstruct surface that did not exist at the current time.

The structured dynamics themselves use fixed scaffold, front, and interface token budgets rather than scaling with full mesh resolution. Growth is therefore represented through the geometric content and organization of the latent state, together with the variable-cardinality geometry that state supports, rather than by increasing the cardinality of the recurrent latent tensor.

\begin{table*}[t]
\footnotesize
\centering
\caption{FOLIAGE preserves global and local geometric identity and aligns heterogeneous sensing modalities. For retrieval, we report image (I)$\rightarrow$points (P), zero-shot I$\rightarrow$P, and zero-shot P$\rightarrow$I.}
\label{tab:t1-geometry}
\setlength{\tabcolsep}{2pt}
\renewcommand{\arraystretch}{1}

\begin{tabular}{@{} c @{\hspace{2pt}} c @{\hspace{2pt}} c @{}}

\begin{tabular}{l r}
\multicolumn{2}{l}{\textbf{Topology Classification}} \\
\multicolumn{2}{r}{$\%$Accuracy$\uparrow$} \\
\midrule
MeshCNN \cite{hanocka2019meshcnn} & 88 \\
DiffusionNet \cite{Sharp2022_DiffusionNet} & 92 \\
\cite{nishikawa2023adaptive} & 94 \\
ETNN \cite{battiloro2023etnn} & 94 \\
\textbf{Ours} & \textbf{98}
\end{tabular}
&
\begin{tabular}{l r}
\multicolumn{2}{l}{\textbf{Dense Correspondence}} \\
\multicolumn{2}{r}{Geodesic Err.$\downarrow$} \\
\midrule
ZoomOut \cite{melzi2019zoomout} & 4.2 \\
DiffusionNet \cite{Sharp2022_DiffusionNet} & 3.8 \\
G-MSM \cite{eisenberger2023g} & 3.6 \\
\cite{cao2024spectral} & 3.2 \\
\textbf{Ours} & \textbf{2.6}
\end{tabular}
&
\begin{tabular}{l r}
\multicolumn{2}{l}{\textbf{Cross-Modal Retrieval}} \\
\multicolumn{2}{r}{mAP@100 (\%)$\uparrow$} \\
\midrule
CrossPoint \cite{zhang2021crosspoint} & 42/18/16 \\
CLIP2Point \cite{zhang2023clip2point} & 43/20/19 \\
ULIP-2 \cite{li2023ulip2} & 46/23/21 \\
PointCLIPv2 \cite{zhu2022pointclip} & 48/22/23 \\
\textbf{Ours} & \textbf{64/39/37}
\end{tabular}

\end{tabular}
\end{table*}

\begin{figure*}[t]
\centering
\setlength{\tabcolsep}{2pt}
\renewcommand{\arraystretch}{1.1}

\begin{tabular}{ccccccc}
    \includegraphics[width=0.13\textwidth]{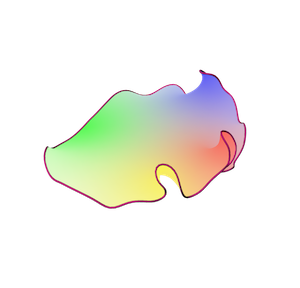} &
    \includegraphics[width=0.13\textwidth]{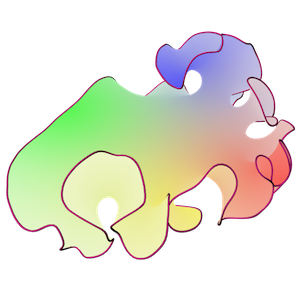} &
    \includegraphics[width=0.13\textwidth]{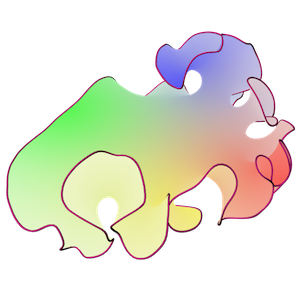} &
    \includegraphics[width=0.13\textwidth]{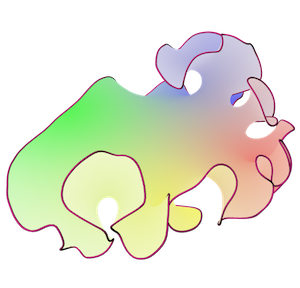} &
    \includegraphics[width=0.13\textwidth]{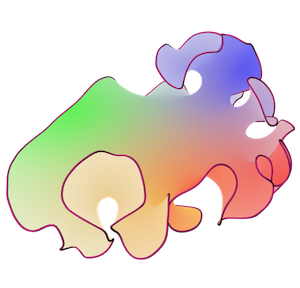} &
    \includegraphics[width=0.13\textwidth]{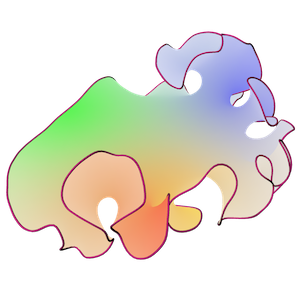} &
    \includegraphics[width=0.13\textwidth]{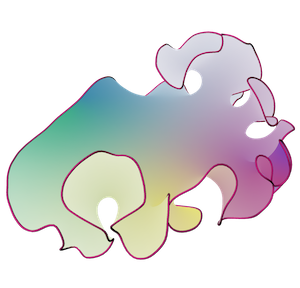} \\
    \small Source &
    \small Target (GT) &
    \small Ours &
    \small \cite{cao2024spectral} &
    \small DiffusionNet &
    \small G-MSM &
    \small ZoomOut
\end{tabular}

\caption{FOLIAGE accurately tracks features in dense correspondence tasks as baselines degrade under significant surface morphology changes caused by accretive growth.}
\label{fig:qual-corr}
\vspace{-0.4cm}
\end{figure*}

\begin{table*}[t]
\footnotesize
\centering
\caption{FOLIAGE captures hidden causes and models counterfactual futures}
\label{tab:t2-world-modeling}
\setlength{\tabcolsep}{2pt}
\renewcommand{\arraystretch}{1}

\begin{tabular}{@{} c @{\hspace{2pt}} c @{\hspace{2pt}} c @{}}

\begin{tabular}{l r}
\multicolumn{2}{l}{\textbf{Inverse Material Regression}} \\
\multicolumn{2}{r}{MAE$(\times10^3)\downarrow$} \\
\midrule
\cite{Kairanda2024_NeuralClothSim} & 60 \\
DiffPD \cite{hu2021diffpd} & 58 \\
BDP \cite{gong2024bayesian} & 55 \\
DiffCloth \cite{li2022diffcloth} & 53 \\
\textbf{Ours} & \textbf{32}
\end{tabular}
&
\begin{tabular}{l r}
\multicolumn{2}{l}{\textbf{Growth-Stage Recognition}} \\
\multicolumn{2}{r}{$\%$Balanced Acc.$\uparrow$} \\
\midrule
SVFormer \cite{chen2023svformer} & 67 \\
VideoMAE-v2 \cite{wang2023videomaev2} & 68 \\
TimeSformer \cite{bertasius2021timesformer} & 69 \\
VideoMamba \cite{li2024videomamba} & 71 \\
\textbf{Ours} & \textbf{82}
\end{tabular}
&
\begin{tabular}{l r}
\multicolumn{2}{l}{\textbf{Mesh Forecasting (direct 5-step)}} \\
\multicolumn{2}{r}{Chamfer$(\times10^3)\downarrow$ / Vertex Drift$\downarrow$} \\
\midrule
MeshGraphNets \cite{Pfaff2021_MeshGraphNets} & 65/4133 \\
CaDeX \cite{Lei2022CaDeX} & 52/2261 \\
MeshGPT-solo \cite{siddiqui2024meshgpt} & 45/2540 \\
Motion2VecSets \cite{Cao2024Motion2VecSets} & 38/1721 \\
\textbf{Ours} & \textbf{26/915}
\end{tabular}

\end{tabular}
\vskip -0.3cm
\end{table*}

\begin{figure*}[t]
    \centering
    \setlength{\tabcolsep}{2pt} 
    \renewcommand{\arraystretch}{1.0}

    \begin{tabular}{cccccc}
        \includegraphics[width=0.15\textwidth]{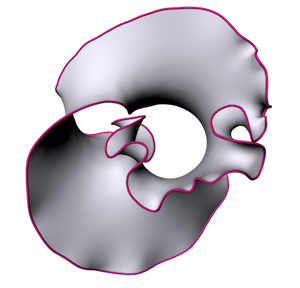} &
        \includegraphics[width=0.15\textwidth]{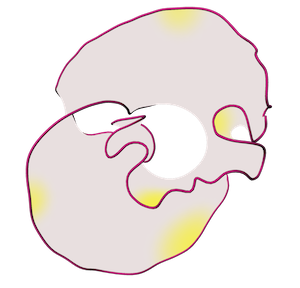} &
        \includegraphics[width=0.15\textwidth]{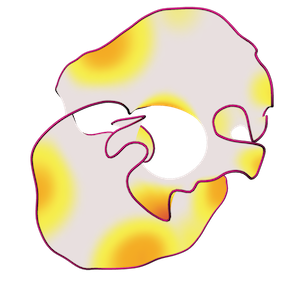} &
        \includegraphics[width=0.15\textwidth]{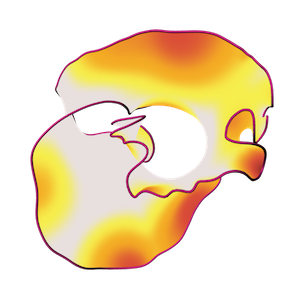} &
        \includegraphics[width=0.15\textwidth]{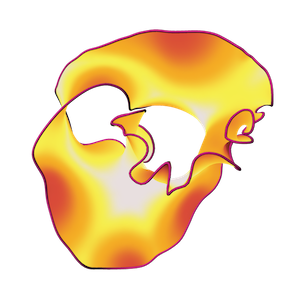} &
        \includegraphics[width=0.15\textwidth]{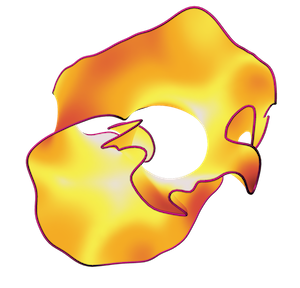} \\

        \small Ground Truth &
        \small Ours &
        \small Motion2VecSets &
        \small MeshGPT-solo &
        \small CaDeX &
        \small MeshGraphNets \\
    \end{tabular}

    \caption{Mesh predictions over $10$ time steps under control actions. Higher errors = more red. FOLIAGE's predictions closely resemble the ground truth while baselines' rapidly deviate.}
    \label{fig:qual-pred}
    \vspace{-0.4cm}
\end{figure*}

\begin{table*}[t]
\footnotesize
\centering
\caption{Component ablations and matched allocation controls for FOLIAGE. Left and middle report topology classification accuracy (A), inverse-material MAE $\times10^3$ (M), direct mesh-forecasting Chamfer $\times10^3$ (C), and retrieval mAP (\%) (m). Right isolates state allocation and dynamics under a matched 448-token, approximately 43M-parameter budget; C5 is direct 5-step Chamfer and C40 is recursive 40-step Chamfer, both $\times10^3$.}
\label{tab:t3-ablations}
\setlength{\tabcolsep}{2pt}
\renewcommand{\arraystretch}{1}

\begin{tabular}{@{} c @{\hspace{2pt}} c @{\hspace{2pt}} c @{}}

\begin{tabular}{l c c c c}
\multicolumn{5}{l}{\textbf{Effect of Growth-adaptive state}} \\
\midrule
Variant & A$\uparrow$ & M$\downarrow$ & C$\downarrow$ & m$\uparrow$ \\
\midrule
FOLIAGE & \textbf{97.7} & \textbf{32} & \textbf{25.7} & \textbf{64.2} \\
w/o GSA & 95.8 & 35 & 28.0 & 63.0 \\
random front & 95.6 & 36 & 28.7 & 62.4 \\
front-only & 94.9 & 34 & 29.4 & 60.4 \\
w/o interface & 96.4 & 33 & 26.6 & 63.6 \\
\end{tabular}
&
\begin{tabular}{l c c c c}
\multicolumn{5}{l}{\textbf{Effect of interface and supervision}} \\
\midrule
Variant & A$\uparrow$ & M$\downarrow$ & C$\downarrow$ & m$\uparrow$ \\
\midrule
FOLIAGE & \textbf{97.7} & \textbf{32} & \textbf{25.7} & \textbf{64.2} \\
w/o corr. & 96.1 & 39 & 31.8 & 50.2 \\
w/o masking & 96.1 & 37 & 29.8 & 55.8 \\
w/o age feat. & 95.2 & 40 & 34.4 & 62.1 \\
w/o energies & 96.3 & 50 & 28.8 & 63.3 \\
w/o nrg. gate & 96.4 & 46 & 28.3 & 63.1 \\
mean pool & 95.5 & 38 & 32.0 & 57.3 \\
\end{tabular}
&
\begin{tabular}{l c c c c}
\multicolumn{5}{l}{\textbf{Matched allocation \& dynamics}} \\
\midrule
Method & Params & GFLOPs & C5$\downarrow$ & C40$\downarrow$ \\
\midrule
Uniform FPS       & 42.8M & 17.67 & 28.2 & 62.7 \\
Learned queries   & 43.0M & 17.72 & 27.8 & 59.8 \\
Generic pruning   & 43.1M & 17.81 & 27.1 & 53.6 \\
GSA + flat dyn.  & 43.0M & 17.75 & 26.4 & 47.9 \\
\textbf{FOLIAGE}  & 43.0M & 17.82 & \textbf{25.7} & \textbf{41.2} \\
\end{tabular}

\end{tabular}
\vskip -0.3cm
\end{table*}

\begin{table*}[t]
\footnotesize
\centering
\caption{FOLIAGE is robust under partial sensing and imperfect correspondences. (L) We vary modality availability: Rich (all), Typical (RGB + point clouds, incomplete meshes), Sparse (one modality), and Noisy (sensor noise + random modality dropout) and report the best-performing models. (R) We increase correspondence corruption: Corr-noise (local rewiring + edge drop from the multimodal correspondence graph), Occl. (RGB masking with linked 3D tokens), Drift (camera jitter), and SLAM (pose re-estimation).}
\label{tab:t4-deployment}
\setlength{\tabcolsep}{2pt}
\renewcommand{\arraystretch}{1}

\begin{tabular}{@{} c @{\hspace{5pt}} c @{}}

\begin{tabular}{l c c c c}
\multicolumn{5}{l}{\textbf{Sensor-Subset Robustness}} \\
\multicolumn{5}{r}{Growth-stage recognition Balanced Acc.$\uparrow$} \\
\midrule
Method & Rich & Typical & Sparse & Noisy \\
\midrule
VideoMAE-v2 & 0.74 & 0.68 & -- & 0.62 \\
PiMAE \cite{chen2023pimae} & 0.76 & 0.70 & 0.67 & 0.63 \\
ULIP-2 & 0.78 & 0.72 & 0.71 & 0.66 \\
CLIP2Point & 0.74 & 0.65 & 0.68 & 0.60 \\
\textbf{Ours} & \textbf{0.80} & \textbf{0.78} & \textbf{0.74} & \textbf{0.74} \\
\end{tabular}
&
\begin{tabular}{l c c c c c}
\multicolumn{6}{l}{\textbf{Correspondence Robustness}} \\
\multicolumn{6}{r}{Cross-modal retrieval mAP@100$\uparrow$} \\
\midrule
Method & Clean & Corr. & Occl. & Drift & SLAM \\
\midrule
\textbf{FOLIAGE} & \textbf{0.60} & \textbf{0.56} & \textbf{0.51} & \textbf{0.52} & \textbf{0.59} \\
FOLIAGE w/o corr. & 0.46 & 0.45 & 0.43 & 0.44 & 0.46 \\
ULIP-2 & 0.46 & 0.49 & 0.40 & 0.41 & 0.42 \\
PointCLIPv2 & 0.48 & 0.47 & 0.39 & 0.40 & 0.41 \\
CLIP2Point & 0.43 & 0.45 & 0.37 & 0.38 & 0.39 \\
CrossPoint & 0.42 & 0.43 & 0.35 & 0.36 & 0.37 \\
\end{tabular}

\end{tabular}
\vspace{-0.6cm}
\end{table*}

\begin{table*}[t]
\centering
\footnotesize
\caption{Representation transfer beyond SURF-GARDEN. FOLIAGE transfers to \textbf{real plant datasets} GrowliFlower \cite{Kierdorf2023GrowliFlower}, Pheno4D \cite{Schunck2021Pheno4D}, and Crops3D \cite{Zhu2024Crops3D}, and to \textbf{dynamic human-body datasets} D-FAUST \cite{Bogo2017_DFAUST} and CAPE \cite{Ma2020CAPE}. ``Frozen'' keeps the encoder fixed; ``Few-shot'' trains a linear probe using 10 labeled sequences per class; ``Light-ft'' additionally fine-tunes the last encoder block.}
\label{tab:t5-transfer}
\setlength{\tabcolsep}{4pt}
\renewcommand{\arraystretch}{1.05}

\begin{minipage}[t]{0.34\textwidth}
\centering
\begin{tabular}{lc}
\toprule
\textbf{Method} & \textbf{GrowliFlower} \\
\multicolumn{2}{r}{Growth stage recognition Balanced Acc. $\uparrow$} \\
\midrule
FOLIAGE (Frozen)      & 0.72 \\
FOLIAGE (Few-shot)    & 0.75 \\
FOLIAGE (Light-ft)    & \textbf{0.77} \\
\midrule
VideoMAE-v2           & 0.70 \\
TimeSformer           & 0.68 \\
VideoMamba            & 0.69 \\
SVFormer              & 0.66 \\
\bottomrule
\end{tabular}
\end{minipage}
\hfill
\begin{minipage}[t]{0.64\textwidth}
\centering
\begin{tabular}{lcccc}
\toprule
\textbf{Method} & \textbf{Pheno4D} & \textbf{Crops3D} & \textbf{D-FAUST} & \textbf{CAPE} \\
 & \multicolumn{4}{c}{Dense correspondence Geodesic. Err. $\downarrow$} \\
\midrule
FOLIAGE (Frozen)      & 3.55 & \textbf{4.45} & 3.6 & 4.4 \\
FOLIAGE (Light-ft)    & \textbf{3.25} &  --  & \textbf{3.1} & \textbf{3.8} \\
\midrule
DiffusionNet          & 3.90 & 4.80 & 3.8 & 4.7 \\
G-MSM                 & 4.60 & 5.50 & 4.5 & 5.4 \\
\cite{cao2024spectral}& 5.05 & 5.95 & 4.9 & 5.8 \\
ZoomOut               & 5.70 & 6.70 & 5.6 & 6.6 \\
\bottomrule
\end{tabular}
\end{minipage}
\vspace{-0.4cm}
\end{table*}

\begin{table}[t]
\centering
\footnotesize
\caption{Passive future-geometry forecasting on real temporal plant scans. Chamfer distance is bounding-box normalized and reported as $\times10^3$; brackets are 95\% bootstrap intervals over held-out plant trajectories after averaging three fine-tuning seeds. Persistence copies the current surface, constant velocity extrapolates motion from the preceding scans, and Uniform AR is a matched variable-cardinality autoregressive model. Splits are by physical plant, and all learned models are lightly fine-tuned on the corresponding training plants.}
\label{tab:t6-real-forecast}
\setlength{\tabcolsep}{4pt}
\renewcommand{\arraystretch}{1.05}
\begin{tabular}{lccc}
\toprule
Method &
Pheno4D Maize, 2-day$\downarrow$ &
Pheno4D Tomato, 2-day$\downarrow$ &
Crops3D Cabbage, 2-week$\downarrow$ \\
\midrule
Persistence &
32.4 [30.4,34.4] &
39.8 [37.2,42.4] &
50.8 [47.4,54.2] \\
Constant velocity &
27.6 [25.8,29.4] &
33.6 [31.3,35.9] &
43.2 [40.2,46.2] \\
Uniform AR, light-ft &
23.1 [21.6,24.6] &
28.6 [26.6,30.6] &
37.2 [34.6,39.8] \\
\textbf{FOLIAGE, light-ft} &
\textbf{19.6 [18.3,20.9]} &
\textbf{24.2 [22.5,25.9]} &
\textbf{31.3 [29.1,33.5]} \\
\bottomrule
\end{tabular}
\vskip -0.3cm
\end{table}

\section{Results}

We evaluate whether FOLIAGE learns a deployable geometric state for growing surfaces. Such a state should preserve global geometry, track identity through growth, ground heterogeneous sensors, encode hidden material causes, and support future prediction under material actions. SURF-GARDEN provides the controlled setting needed to test these properties through matched counterfactual branches, dense temporal and cross-modal correspondences, hidden physical signals, and complete surface geometry. SURF-BENCH then probes the learned state through capability-focused tasks: geometry and multimodal grounding in Sec.~\ref{sec:results-geometry}, hidden-cause modeling and action-conditioned forecasting in Sec.~\ref{sec:results-world-model}, architectural ablations in Sec.~\ref{sec:results-ablation}, and robustness and real-world generalization in Sec.~\ref{sec:results-deployment}.

\subsection{Does FOLIAGE preserve geometry and identity for growing surfaces?} \label{sec:results-geometry}

If FOLIAGE learns geometry as state, it should preserve both global surface structure and local identity as new material appears. We test this in two ways: topology classification tests whether the latent retains global geometric invariants; dense correspondence tests whether it tracks the same physical or semantic regions through growth. These are state tests: a model that predicts plausible local shape but loses topology or correspondence has not learned a reliable physical state.

In Tab.~\ref{tab:t1-geometry}, FOLIAGE significantly improves $\%$ topology accuracy from $94$ to $98$ and reduces dense-correspondence geodesic error from $3.2$ to $2.6$. Fig.~\ref{fig:qual-corr} gives the qualitative counterpart: FOLIAGE tracks features through growth-induced morphology changes more consistently than static correspondence baselines. Together, these results show that the learned state preserves both what the surface is globally and how its regions persist locally through growth.

\textbf{Multimodal grounding under partial sensors.}
If FOLIAGE builds one geometric state from partial sensors, RGB and 3D observations of the same growing surface should align in the learned space. We test this with image--point retrieval: the question is whether heterogeneous observations are grounded to the same physical surface state.
In Tab.~\ref{tab:t1-geometry}, FOLIAGE substantially improves supervised image--point retrieval; in zero-shot retrieval, FOLIAGE also remains ahead in both image-to-point and point-to-image directions. These gains indicate that correspondence-aware geometric fusion produces a shared state across modalities, rather than separate image and 3D embeddings that only work under task-specific supervision.

\subsection{Is FOLIAGE a good world model? Hidden causes and counterfactual futures} \label{sec:results-world-model}

If FOLIAGE learns a physical state for growing geometry, the state should encode hidden causes of future morphology and support different futures under different material actions. We test this in three ways: material estimation probes whether the latent captures physical factors that drive growth; stage recognition asks whether it captures where the process is in its trajectory; forecasting tests whether the state can be rolled forward under action-conditioned dynamics.
In Tab.~\ref{tab:t2-world-modeling}, FOLIAGE reduces inverse-material mean absolute error and 5-step mesh-forecasting Chamfer distance, while also improving growth-stage recognition and vertex drift. Fig.~\ref{fig:qual-pred} gives the qualitative counterpart: FOLIAGE better preserves global morphology and active-front detail at the queried future horizon. Together, these results show that the learned state contains both information about hidden physical causes and information needed to predict future geometry.

\textbf{Does the action actually select the future?}
We additionally evaluate counterfactual sibling branches that share the same observed prefix but evolve under different material controls. Given the queried action, FOLIAGE retrieves the corresponding ground-truth sibling future in $90.4\%$ of cases and obtains a predicted-to-true branch-separation ratio of $0.95$ ($1$ is ideal). Moving the action into the state encoder instead reduces these values to $78.2\%$ and $0.84$. This supports the intended interface: the inferred current state describes the observed surface, while the transition selects among alternative action-conditioned futures.

\subsection{Why Does FOLIAGE Work?} \label{sec:results-ablation}

FOLIAGE’s gains should come from design choices matched to growing geometry, not from undifferentiated model capacity. In Tab.~\ref{tab:t3-ablations}, we test this by ablating growth-adaptive state allocation, then by removing the interfaces that ground sensors, localize growth, and use train-only physical signals.

\textbf{Effect of GSA.} Removing GSA and reverting to a flat pooled state hurts forecasting and material inference. Random front selection underperforms despite using a similar token budget, showing that activity-guided allocation matters. Front-only updates lose topology and retrieval performance, showing that local growth must remain anchored to mature geometry. Removing interface tokens weakens forecast consistency. Full FOLIAGE performs best because growth is local but not isolated: active regions must be updated in the context of the mature surface they remain attached to.

\textbf{Other design choices.} Further ablations show that they fail in interpretable ways. Removing correspondence fusion mainly hurts retrieval; removing age/activity features dents growth-sensitive forecasting; removing train-only energies or only energy-gated message passing mainly limits material inference. Each component affects the capability it was introduced to support.

\textbf{Matched allocation and dynamics.}
The direct GSA ablation establishes that growth-aware state allocation helps, but does not by itself distinguish growth-specific allocation from generic adaptive token selection. We therefore compare same-budget alternatives with identical encoders and readouts and matched transition computation. The complete scaffold/front/interface-specific dynamics in FOLIAGE outperforms the alternatives. The gains are greater on C40 than on C5, suggesting that the value of structured allocation becomes clearest under recursive prediction, where local allocation and update errors compound over time.

\textbf{When is growth-adaptive allocation most useful?}
GSA is designed for spatially nonuniform change. As active coverage becomes global, its advantage over uniform allocation narrows and activity recall falls, with errors concentrating on missed high-change regions (Appendix~\ref{sec:distributed-growth}). The same ordering persists through the complete 350-frame rollout (Appendix~\ref{sec:full-rollout}). These results characterize GSA's intended regime: adaptive allocation helps most when future change is spatially heterogeneous.

\subsection{Does FOLIAGE generalize under imperfect sensing and beyond simulation?}\label{sec:results-deployment}

\textbf{Robustness under realistic sensing conditions.}
Tab.~\ref{tab:t4-deployment} tests missing modalities and corrupted cross-modal correspondences. FOLIAGE leads across rich, typical, sparse, and noisy sensing, and remains strongest under correspondence noise, occlusion, and camera drift. Performance degrades smoothly as sensing weakens, while the correspondence-free ablation remains usable but lower. Thus, correspondence-aware fusion improves geometric grounding without requiring ideal links, while masking and modality dropout reduce dependence on any single sensor.

We then test whether correspondence-aware fusion requires ideal links (Tab.~\ref{tab:t4-deployment}). FOLIAGE remains strongest under correspondence noise, occlusion, and camera drift, and recovers when pose refinement is available. The correspondence-free ablation remains usable but lower. This is expected: correspondences improve grounding when approximate associations exist, but they are not an oracle dependency.
These robustness tests show that FOLIAGE’s state remains usable under realistic sensing stress like missing modalities, noisy links, and correspondence drift.


\textbf{Real-world and cross-domain representation transfer.}
If FOLIAGE learns reusable growing-geometry features rather than simulator-specific shortcuts, its representation should remain useful outside SURF-GARDEN.
Tab.~\ref{tab:t5-transfer} tests whether features learned in SURF-GARDEN remain useful on independently collected data. FOLIAGE improves real-plant growth-stage recognition and transfers to dense correspondence on plant and dynamic human-body surfaces. These experiments test representation transfer rather than the learned transition itself.
Hence, for \textbf{real-world future-geometry forecasting,} we execute the transition model on temporal 3D plant scans (Tab.~\ref{tab:t6-real-forecast}). With no intervention labels, FOLIAGE uses its passive null-action token and improves future-geometry prediction across all three settings, including over a matched Uniform AR baseline.

\section{Scope and Limitations}

FOLIAGE targets \emph{accretive thin surfaces}, rather than arbitrary volumetric growth or unrestricted topology changes such as tearing and merging. Within this regime, we vary topology, material response, sensing, growth localization, rollout horizon, and controls, and additionally evaluate passive future prediction on real plant scans. Intervention-conditioned counterfactuals are evaluated in simulation because available real datasets do not provide matched futures under controlled material actions, leaving real closed-loop control outside the demonstrated scope. Performance also degrades under globally distributed activity, sufficiently poor sensing, and long open-loop rollouts, as quantified by our stress tests. Richer real 3D capture, online state refresh, real intervention datasets, and extensions to more general growth processes are natural directions beyond the setting studied here.

\section{Conclusion}

We introduced FOLIAGE, a geometry-centered latent world model for growing surfaces. FOLIAGE organizes a fixed state budget into mature scaffold, future-active front, and interface tokens, then applies role-specific dynamics while using privileged physics only during training. Across SURF-BENCH, matched allocation controls, robustness tests, and real plant forecasting, this growth-adaptive state improves geometric representation and future prediction. Together, the controlled counterfactual experiments and real passive forecasting show that the learned state remains useful both for reasoning about alternative growth futures and for predicting geometry beyond the simulator. These results support a simple principle for growing systems: representation capacity should follow where new physical structure is emerging while preserving compact global context. 

\bibliographystyle{plainnat}
\bibliography{main}

\appendix

\newpage
\section{Appendix}

In this appendix, we provide implementation details, dataset details, and further experimental validation.

\begin{figure*}[ht]
    \centering
    \setlength{\tabcolsep}{2pt} 
    \renewcommand{\arraystretch}{1.2} 
    \begin{tabular}{ccc}
        \includegraphics[width=0.28\textwidth]{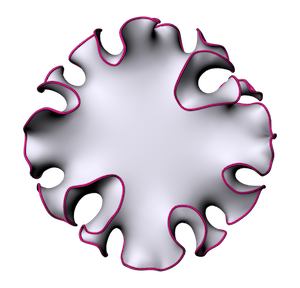} &
        \includegraphics[width=0.28\textwidth]{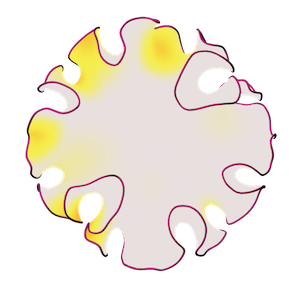} &
        \includegraphics[width=0.28\textwidth]{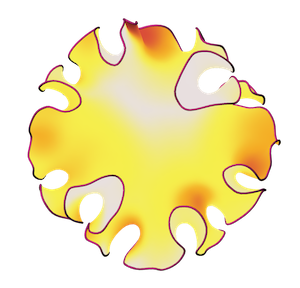} \\
        \small Ground Truth &
        \small Ours &
        \small Motion2VecSets \cite{Cao2024Motion2VecSets} \\
        \includegraphics[width=0.28\textwidth]{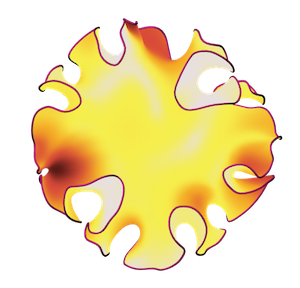} &
        \includegraphics[width=0.28\textwidth]{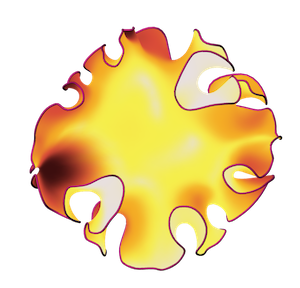} &
        \includegraphics[width=0.28\textwidth]{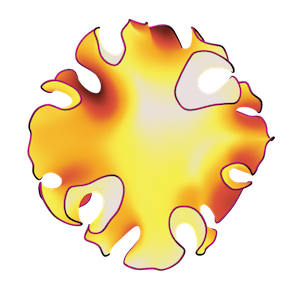} \\
        \small MeshGPT-solo \cite{siddiqui2024meshgpt} &
        \small CaDeX \cite{Lei2022CaDeX} &
        \small MeshGraphNets \cite{Pfaff2021_MeshGraphNets} \\
    \end{tabular}
    \caption{Mesh predictions on SURF-BENCH. $\Delta t = 18$, action = $[0.1, 0.3, 0.01]$.} \label{fig:t4-1}
\end{figure*} 

\begin{figure*}[ht]
    \centering
    \setlength{\tabcolsep}{2pt} 
    \renewcommand{\arraystretch}{1.2} 
    \begin{tabular}{ccc}
        \includegraphics[width=0.28\textwidth]{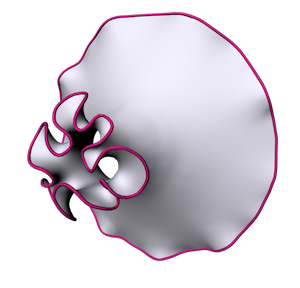} &
        \includegraphics[width=0.28\textwidth]{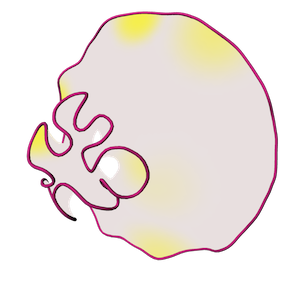} &
        \includegraphics[width=0.28\textwidth]{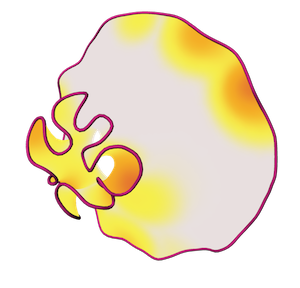} \\
        \small Ground Truth &
        \small Ours &
        \small Motion2VecSets \cite{Cao2024Motion2VecSets} \\
        \includegraphics[width=0.28\textwidth]{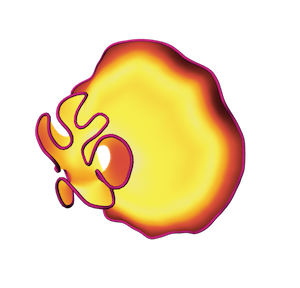} &
        \includegraphics[width=0.28\textwidth]{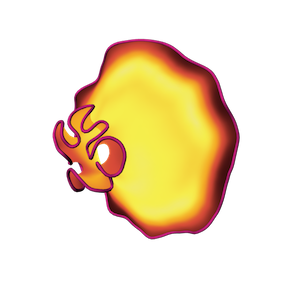} &
        \includegraphics[width=0.28\textwidth]{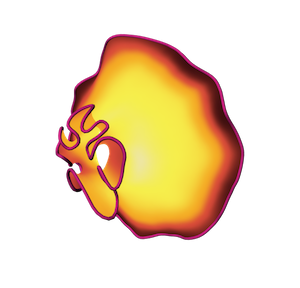} \\
        \small MeshGPT-solo \cite{siddiqui2024meshgpt} &
        \small CaDeX \cite{Lei2022CaDeX} &
        \small MeshGraphNets \cite{Pfaff2021_MeshGraphNets} \\
    \end{tabular}
    \caption{Mesh predictions on SURF-BENCH. $\Delta t = 8$, null action.} \label{fig:t4-2} 
\end{figure*} 

\begin{figure*}[ht]
    \centering
    \setlength{\tabcolsep}{2pt} 
    \renewcommand{\arraystretch}{1.2} 
    \begin{tabular}{ccc}
        \includegraphics[width=0.28\textwidth]{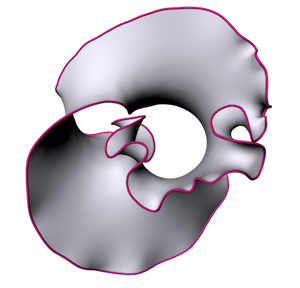} &
        \includegraphics[width=0.28\textwidth]{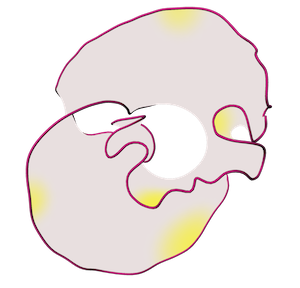} &
        \includegraphics[width=0.28\textwidth]{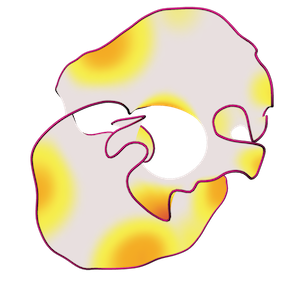} \\
        \small Ground Truth &
        \small Ours &
        \small Motion2VecSets \cite{Cao2024Motion2VecSets} \\
        \includegraphics[width=0.28\textwidth]{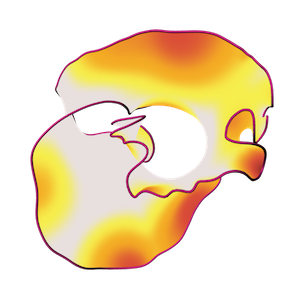} &
        \includegraphics[width=0.28\textwidth]{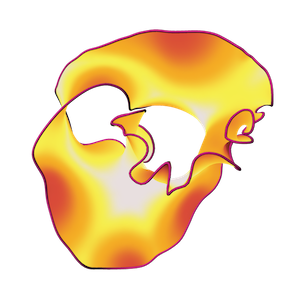} &
        \includegraphics[width=0.28\textwidth]{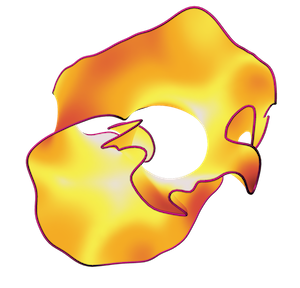} \\
        \small MeshGPT-solo \cite{siddiqui2024meshgpt} &
        \small CaDeX \cite{Lei2022CaDeX} &
        \small MeshGraphNets \cite{Pfaff2021_MeshGraphNets} \\
    \end{tabular}
    \caption{Mesh predictions on SURF-BENCH. $\Delta t = 10$, action = $[0.05, 0.2, 0.12]$.} \label{fig:t4-3}
\end{figure*} 

\section{Sensor Tokenization and Correspondence Graph}

This section provides implementation details for the geometric tokenization.

\paragraph{Shared token dimension.}
All modality tokens are projected to a shared dimension ($d = 768$). Unless otherwise stated, FOLIAGE uses the same token dimension for image, point, mesh, scaffold, front, boundary, action, and horizon tokens.

\paragraph{Image tokens.}
Each RGB image $I_t$ is divided into non-overlapping $16 \times 16$ patches and encoded with a lightweight ViT \cite{VIT} backbone. Patch features are linearly projected to dimension $d$, producing image tokens
\[
T_I = \{ u^I_k \}_{k=1}^{N_I}.
\]
Image tokens carry a modality embedding and camera/view embedding. They do not carry intrinsic age; when image-to-mesh correspondence is available, age and growth-activity supervision are inherited from the corresponding mesh element.

\paragraph{Point tokens.}
Each point cloud $P_t$ is downsampled to $2{,}048$ points using farthest-point sampling. A local point transformer groups these into $N_P = 512$ point tokens,
\[
T_P = \{ u^P_k \}_{k=1}^{N_P}.
\]
Point features include 3D position, estimated normal, local density, and relative offsets within the local point neighborhood. As with image tokens, point tokens inherit age and activity targets through nearby mesh correspondences when available. We use PointNeXt-L \cite{qian2022pointnext} in a two-level PointNet++ \cite{ptpp} hierarchy. Training augmentations include random point dropout, jitter, and global rotations to mimic LiDAR noise and sparsity.

\paragraph{Mesh tokens.}
The mesh $M_t = (V_t, E_t, F_t)$ can contain many more vertices than the model can process directly. We therefore form a mesh-anchor token set $T_M$ with a budget of $N_M = 2048$. Young vertices are prioritized up to half of the budget, and the remaining anchors are selected by geodesic farthest-point sampling over the surface. Each mesh token contains position, normal, local area, curvature estimate, vertex degree, and a birth-time-derived age feature. If vertex $v$ first appears at time $\tau(v)$, we use normalized age
\[
\alpha_t(v) = \mathrm{clip}\!\left( \frac{t - \tau(v)}{\Delta_{\mathrm{age}}}, 0, 1 \right),
\]
with $\Delta_{\mathrm{age}} = 50$, and a recency feature
\[
r_t(v) = \exp\!\left( -\frac{\alpha_t(v)}{\tau_{\mathrm{rec}}} \right),
\]
with $\tau_{\mathrm{rec}} = 0.2$. Age is used as an observable proxy for material maturity. We use two mesh diffusion layers (DiffusionNet \cite{Sharp2022_DiffusionNet}) to capture local geometry, followed by two learned-step graph ODEs (GRAND++ \cite{Chamberlain2021_GRAND, Thorpe2022_GRANDplusplus}) handles evolving connectivity.

\paragraph{Correspondence graph.}
Fused tokens interact through a sparse heterogeneous graph $G_t = (U_t, \mathcal{E}_t)$. The edge set includes:
\begin{enumerate}
    \item \textbf{Image-to-mesh edges:} each image token connects to visible mesh elements projected into the corresponding patch. Edge features include barycentric weights, depth, visibility confidence, and view direction.
    
    \item \textbf{Point-to-mesh edges:} each point token connects to its three nearest mesh anchors. Edge features include Euclidean distance, normal agreement, and inverse-distance confidence.
    
    \item \textbf{Mesh-to-mesh edges:} mesh anchors connect through one-hop mesh adjacency and a small set of geodesic nearest neighbors.
    
    \item \textbf{Within-modality local edges:} point tokens and image tokens retain local neighborhood edges within their modality.
\end{enumerate}

Fusion uses type-aware sparse attention:
\[
\tilde{u}_i
=
u_i
+
\sum_{j \in \mathcal{N}(i)}
\mathrm{Attn}_{\mathrm{type}(i),\,\mathrm{type}(j)}
\big( u_i, u_j, b_{ij} \big),
\]
where $b_{ij}$ contains the edge features above. We use two sparse fusion layers in the main model.

\paragraph{Age propagation to non-mesh tokens.}
Image and point tokens inherit an age proxy from neighboring mesh anchors through correspondence edges. If multiple mesh anchors are associated with a token, we use a conservative minimum-age rule, assigning the token the youngest nearby age. Tokens without valid mesh correspondence receive a learned unknown-age embedding and are excluded from age-supervised losses.

\paragraph{Structured masking.}
During training, FOLIAGE applies three masking schemes: random token masking within each active modality, paired masking of tokens connected by correspondence edges, and modality-level dropout. We use modality dropout probability $0.3$. Paired masking prevents the model from relying on trivial one-to-one correspondence copying and improves robustness when local registrations are missing or corrupted.

\section{Growth-Adaptive State Allocation Implementation}

This section provides implementation details for GSA, the state-allocation module.

\paragraph{Default token counts.}
Unless otherwise stated, FOLIAGE uses $K_m = 128$ mature scaffold tokens, $K_f = 256$ active growth-front tokens, and $K_b = 64$ interface tokens. These values are chosen to keep the dynamics state substantially smaller than the full mesh while retaining enough active tokens for local growth.

\paragraph{Activity score.}
For each fused token $u_i$, the activity score is
\[
g_i =
\sigma \left(
f_\psi\big([u_i, r_i, c_i, \rho_i, m_i]\big)
\right),
\]
where $r_i$ is the recency feature, $c_i$ contains local curvature and normal-variation features, $\rho_i$ is average correspondence confidence, and $m_i$ is a modality mask. $f_\psi$ is a two-layer MLP with hidden dimension $d/2$, GELU activation, and sigmoid output. For non-mesh tokens, $r_i$ and $c_i$ are inherited from corresponding mesh anchors when available; otherwise learned unknown-feature embeddings are used.

\paragraph{Mature scaffold selection.}
Mature scaffold tokens are selected from low-activity or older regions. We sample $K_m$ tokens by weighted farthest-point sampling, with sampling priority
\[
w_i^{\mathrm{mat}} = (1 - g_i)(1 - r_i).
\]
When mesh geodesic distances are available, farthest-point sampling uses geodesic distance; otherwise, it uses Euclidean distance in 3D. The selected scaffold tokens are refined by a graph-attention block over mesh-neighborhood and correspondence edges.

\paragraph{Growth-front selection.}
Growth-front tokens are selected as the top $K_f$ tokens under $g_i$. When mesh observations are available, at least 25\% of the selected front tokens are required to be mesh anchors. This prevents visually salient but geometrically unsupported image patches from dominating the active state. If fewer than $K_f$ valid high-activity tokens are available, the front set is padded with tokens adjacent to selected front tokens in the correspondence graph.

\paragraph{Interface token selection.}
Interface tokens are selected from the transition region between active front and mature scaffold. A token is eligible if it lies within two mesh-graph hops of a selected front token and has at least one graph neighbor outside the front. Eligible tokens are ranked by
\[
g_i(1 - g_i) + \eta \, \mathrm{cut}_i,
\]
where $\mathrm{cut}_i$ counts graph edges crossing between front and non-front regions and $\eta = 0.1$. The top $K_b$ eligible tokens form $S_t^{\mathrm{bdry}}$. If the eligible set is too small, we fill the remaining slots with nearest non-front neighbors of front tokens.

\paragraph{Global summary.}
The global state vector is computed by attention pooling over all fused tokens:
\[
s_t^{\mathrm{global}} = \mathrm{AttnPool}(U_t).
\]
The final structured state is
\[
S_t = \left\{
s_t^{\mathrm{global}},
S_t^{\mathrm{mat}},
S_t^{\mathrm{front}},
S_t^{\mathrm{bdry}}
\right\}.
\]

\paragraph{Ablation variants.}
We evaluate the following GSA variants:
\begin{itemize}
    \item \textbf{Flat / homogeneous FOLIAGE:} replaces the structured state with the previous global-plus-young pooled latent.
    \item \textbf{Random front:} selects $K_f$ front tokens uniformly at random while keeping the same token budget.
    \item \textbf{Front-only:} keeps active front tokens but removes mature scaffold and interface tokens from the dynamics state.
    \item \textbf{Scaffold + front, no interface:} keeps mature and front tokens but removes boundary/interface tokens.
    \item \textbf{Full GSA:} uses mature scaffold, active front, and interface tokens.
\end{itemize}

These variants isolate whether gains come from growth-aware selection, mature scaffold context, and front--scaffold interface modeling rather than from additional capacity alone.

\paragraph{Matched state-allocation controls.}
To separate growth-specific allocation from generic token selection, we additionally compare GSA against four matched alternatives. All variants use 448 non-global state tokens, approximately 43M parameters, the same observation encoders, geometry readout, training data, losses, and comparable transition computation. \textbf{Uniform FPS} allocates the state across the whole observed surface using farthest-point sampling, without an activity-dependent distinction between mature and future-active regions.
\textbf{Learned queries} replaces geometry-selected state tokens with generic learned latent registers.
\textbf{Generic pruning} uses learned adaptive token selection without the scaffold/front/interface decomposition or growth-specific activity target.
\textbf{GSA + flat dynamics} retains growth-aware GSA selection but replaces the role-specific front, interface, and scaffold updates with a homogeneous transition.
Full FOLIAGE combines GSA with the structured dynamics described in Sec.~\ref{sec:method-p3}.

These controls distinguish three possible sources of improvement: additional state capacity, generic adaptive selection, and allocation specialized to predicted growth. Because token count and transition cost are matched, differences primarily reflect how state capacity is assigned and updated rather than how much capacity is available.

\section{Action-Conditioned Dynamics Details}

This section provides implementation details for the dynamics module.

\paragraph{Action encoding.}
Material controls are represented as
\[
a_t = [k_{\mathrm{stretch}}, k_{\mathrm{shear}}, k_{\mathrm{bend}}] \in \mathbb{R}^3.
\]
A two-layer MLP maps $a_t$ to an action token
\[
e_a = \mathrm{MLP}_a(a_t) \in \mathbb{R}^d.
\]
When material controls are unavailable, $e_a$ is replaced by a learned null-action token. During training, the action token is dropped with probability $0.1$ to preserve a passive prediction pathway.

\paragraph{Horizon encoding.}
The horizon $\Delta t$ is encoded using Fourier features:
\[
\phi(\Delta t) =
\Big[
\sin(2^0 \pi \Delta t / T), \cos(2^0 \pi \Delta t / T), \ldots,
\sin(2^{L-1} \pi \Delta t / T), \cos(2^{L-1} \pi \Delta t / T)
\Big].
\]
We use $L = 8$ and $T = 20$. A linear projection maps $\phi(\Delta t)$ to a horizon token $e_\Delta \in \mathbb{R}^d$.

\paragraph{Dynamics input.}
The dynamics transformer receives
\[
Z_t =
\{ s_t^{\mathrm{global}}, e_a, e_\Delta \}
\cup
S_t^{\mathrm{mat}}
\cup
S_t^{\mathrm{front}}
\cup
S_t^{\mathrm{bdry}}.
\]
The structured update is implemented in three stages.

\paragraph{Front update.}
Growth-front tokens receive the highest-capacity update:
\[
\hat{S}_{t+\Delta t}^{\mathrm{front}} =
S_t^{\mathrm{front}} +
F_\theta\left(
S_t^{\mathrm{front}},
\{ s_t^{\mathrm{global}}, e_a, e_\Delta \} \cup S_t^{\mathrm{bdry}}
\right).
\]
$F_\theta$ is a two-layer Transformer block with pre-normalization, local geometric bias, and residual feed-forward updates. Front tokens attend densely to one another and to action, horizon, global, and boundary tokens.

\paragraph{Interface update.}
Interface tokens are updated after the front:
\[
\hat{S}_{t+\Delta t}^{\mathrm{bdry}} =
S_t^{\mathrm{bdry}} +
B_\theta\left(
S_t^{\mathrm{bdry}},
\hat{S}_{t+\Delta t}^{\mathrm{front}},
S_t^{\mathrm{mat}},
s_t^{\mathrm{global}}
\right).
\]
$B_\theta$ is a single cross-attention block. Interface tokens attend to predicted front tokens and nearby scaffold tokens using the graph neighborhoods defined by GSA.

\paragraph{Mature scaffold update.}
Mature scaffold tokens receive a lightweight residual update:
\[
\hat{S}_{t+\Delta t}^{\mathrm{mat}} =
S_t^{\mathrm{mat}} +
\lambda_{\mathrm{mat}} \,
M_\theta\left(
S_t^{\mathrm{mat}},
\hat{S}_{t+\Delta t}^{\mathrm{bdry}},
s_t^{\mathrm{global}}
\right),
\]
with $\lambda_{\mathrm{mat}} = 0.25$. $M_\theta$ is a one-layer graph-attention block restricted to scaffold-neighborhood edges and pooled interface context.

\paragraph{Global update.}
The global token is updated from the predicted structured state:
\[
\hat{s}_{t+\Delta t}^{\mathrm{global}} =
s_t^{\mathrm{global}} +
G_\theta\left(
s_t^{\mathrm{global}},
\mathrm{Pool}(\hat{S}_{t+\Delta t}^{\mathrm{mat}}),
\mathrm{Pool}(\hat{S}_{t+\Delta t}^{\mathrm{front}}),
\mathrm{Pool}(\hat{S}_{t+\Delta t}^{\mathrm{bdry}}),
e_a, e_\Delta
\right).
\]

\paragraph{Multi-horizon training and rollout.}
During training, $\Delta t$ is sampled uniformly from $\{1, \ldots, 20\}$. For direct prediction, the model predicts $\hat{S}_{t+\Delta t}$ in one step using the horizon token. For rollout evaluation, we also recursively apply the one-step transition with $\Delta t = 1$. Direct prediction evaluates whether the state represents requested future geometry; recursive rollout evaluates stability under iteration.

\paragraph{Counterfactual sibling-branch evaluation.}
SURF-GARDEN provides sibling trajectories that share an identical observed prefix and diverge only after different material actions are applied. We use these branches to test whether the predicted state responds to the queried action rather than merely predicting a generic plausible continuation. We report the results in Tab.~\ref{tab:sibling-counterfactual}

For each branch point, \textbf{branch retrieval} asks whether the predicted future is closest to the ground-truth sibling generated by the queried action. We also report a \textbf{separation ratio}, which compares pairwise separation among predicted sibling futures with that among the corresponding ground-truth futures; a value of 1 indicates matching branch separation. C40 is Chamfer $\times10^3$ after 40 recursive one-step transitions.

\begin{table}[t]
\centering
\small
\begin{tabular}{lccc}
\toprule
Method & Retrieval $\uparrow$ & Separation $\uparrow$ & C40 $\downarrow$ \\
\midrule
Null action        & 18.6\% & 0.11 & 66.7 \\
Uniform + action   & 66.8\% & 0.71 & 62.7 \\
Action in encoder  & 78.2\% & 0.84 & 48.8 \\
\textbf{FOLIAGE}   & \textbf{90.4\%} & \textbf{0.95} & \textbf{41.2} \\
\bottomrule
\end{tabular}
\caption{Counterfactual sibling-branch prediction from a shared observed prefix. Keeping the inferred current state independent of the queried action gives the strongest branch identification and separation.}
\label{tab:sibling-counterfactual}
\end{table}

Placing the material action in the observation encoder improves over a generic action-conditioned model but remains worse than introducing the action only in the transition. This supports the intended counterfactual interface: the current state represents the observed surface, while the action selects among possible futures from that state.

\section{Train-Only Physical Guidance Details}

This section provides implementation details for the training-only physics.

\paragraph{Context and target encoders.}
The context encoder $E_{\mathrm{ctx}}$ consumes only observable inputs:
\[
S_t = \mathrm{GSA}(E_{\mathrm{ctx}}(x_t)).
\]
The target encoder $E_{\mathrm{tar}}$ is used only during training and receives observable inputs plus per-vertex physical energies:
\[
S_{t+\Delta t}^{+} =
\mathrm{GSA}^{+}\!\left(
E_{\mathrm{tar}}\big(
x_{t+\Delta t},
e_{t+\Delta t}^{\mathrm{mem}},
e_{t+\Delta t}^{\mathrm{bend}}
\big)
\right).
\]
The target state $S_{t+\Delta t}^{+}$ supervises $\hat{S}_{t+\Delta t}$. The target encoder is updated as an exponential moving average of the context encoder with decay $0.998$. Prediction gradients do not backpropagate through $E_{\mathrm{tar}}$.

\paragraph{Energy-gated message passing.}
In the target encoder, per-vertex membrane and bending energies modulate the first mesh message-passing layer. For a message from vertex $j$ to vertex $i$, we apply
\[
m_{ij}^{+} =
\left(1 + \gamma \tanh(W_e e_j)\right) m_{ij},
\]
where $e_j = [e_j^{\mathrm{mem}}, e_j^{\mathrm{bend}}]$ is the normalized physical energy vector and $\gamma = 0.5$. Energies are detached constants. The gate is bounded by the $\tanh$ nonlinearity and is applied only in the target branch.

\paragraph{Activity supervision.}
The GSA activity score is supervised using both future local change and physical energy. For token $i$, the target activity is
\[
\tilde{g}_i =
\mathrm{Norm}\!\left(
0.5\,\delta_i + 0.5\,\bar{e}_i
\right),
\]
where $\delta_i$ is normalized local displacement between $t$ and $t+\Delta t$, and $\bar{e}_i$ is the normalized sum of membrane and bending energies projected to the token. $\mathrm{Norm}(\cdot)$ rescales values within each surface to $[0,1]$. Image and point tokens inherit $\tilde{g}_i$ from neighboring mesh anchors through correspondence edges. Tokens without valid correspondence are excluded from the activity loss.

The activity loss is
\[
\mathcal{L}_{\mathrm{act}} =
\frac{1}{|\Omega|}
\sum_{i \in \Omega}
\mathrm{BCE}(g_i, \tilde{g}_i),
\]
where $\Omega$ is the set of tokens with valid activity targets.

\paragraph{Energy prediction.}
During training, a lightweight energy head predicts normalized membrane and bending energies from context mesh tokens:
\[
\hat{e}_i = h_{\mathrm{energy}}(u_i).
\]
We supervise this prediction using
\[
\mathcal{L}_{\mathrm{energy}} =
\frac{1}{|\Omega_M|}
\sum_{i \in \Omega_M}
\mathrm{Huber}(\hat{e}_i, e_i).
\]
The energy head is discarded at deployment. Its purpose is to encourage observable features to correlate with hidden stress, not to require physical energies at inference.

\paragraph{Equal privileged-supervision controls.}
To determine whether FOLIAGE's gains arise simply because it receives privileged physical supervision during training, we provide the same physical-energy targets to comparison models under three supervision regimes in Tab.~\ref{tab:equal-physics}. \textbf{No physics} removes privileged-energy supervision. \textbf{Common energy loss} adds the same auxiliary per-vertex energy-prediction objective. \textbf{Matched guidance} additionally uses an EMA target and energy-gated target-side processing analogous to FOLIAGE. Test-time inputs are unchanged in every condition; physical energies are never provided at deployment.

\begin{table}[t]
\centering
\small
\setlength{\tabcolsep}{5pt}
\begin{tabular}{lccc}
\toprule
\multicolumn{4}{c}{Inverse-material MAE $\times10^3$ $\downarrow$} \\
Method & No physics & Common energy loss & Matched guidance \\
\midrule
DiffCloth       & 53.1 & 43.6 & 39.4 \\
Uniform FPS-448 & 52.6 & 42.9 & 35.2 \\
\textbf{FOLIAGE}& 50.1 & 39.2 & \textbf{32.1} \\
\midrule
\multicolumn{4}{c}{Recursive C40 Chamfer $\times10^3$ $\downarrow$} \\
Method & No physics & Common energy loss & Matched guidance \\
\midrule
Motion2VecSets  & 75.4 & 69.0 & 65.8 \\
Uniform FPS-448 & 72.8 & 67.1 & 62.7 \\
\textbf{FOLIAGE}& 54.8 & 47.6 & \textbf{41.2} \\
\bottomrule
\end{tabular}
\caption{Equal privileged-supervision controls. Physical-energy supervision improves each architecture, especially for inverse-material inference, while FOLIAGE retains a substantial forecasting advantage when supervision is matched.}
\label{tab:equal-physics}
\end{table}

Privileged physical labels are therefore useful supervision rather than an exclusive source of FOLIAGE's gains. Their effect is particularly large for hidden-material inference, as expected from the physical quantities they supervise. For recursive forecasting, architecture remains consequential even after supervision is matched: FOLIAGE reaches C40 41.2 compared with 62.7 for matched uniform allocation and 65.8 for Motion2VecSets.

\paragraph{Leakage controls.}
Privileged energies are used only in the target branch and auxiliary training losses. They are detached, bounded through gating, and never provided to the context encoder at inference. The target encoder is EMA-updated rather than directly optimized through privileged gradients from prediction. These controls preserve the intended separation between observation, action, and privileged physics.

\section{Training Objective, Readouts, and Complexity}

This section provides details for the loss, evaluation heads, and computational cost.

\paragraph{Structured latent prediction loss.}
FOLIAGE predicts the future structured state $\hat{S}_{t+\Delta t}$ and matches it to the target state $S_{t+\Delta t}^{+}$. We use one global term and three token-set terms:
\[
\mathcal{L}_{\mathrm{pred}} =
\left\|
\hat{s}_{t+\Delta t}^{\mathrm{global}} -
s_{t+\Delta t}^{+,\mathrm{global}}
\right\|_2^2
+ \lambda_m
D\!\left(
\hat{S}_{t+\Delta t}^{\mathrm{mat}},
S_{t+\Delta t}^{+,\mathrm{mat}}
\right)
+ \lambda_f
D\!\left(
\hat{S}_{t+\Delta t}^{\mathrm{front}},
S_{t+\Delta t}^{+,\mathrm{front}}
\right)
+ \lambda_b
D\!\left(
\hat{S}_{t+\Delta t}^{\mathrm{bdry}},
S_{t+\Delta t}^{+,\mathrm{bdry}}
\right).
\]
We use $\lambda_m = 0.5$, $\lambda_f = 1.0$, and $\lambda_b = 0.75$. The set distance is
\[
D(A, B) =
\frac{1}{|A|}
\sum_{a \in A}
\min_{b \in B} \|a - b\|_2^2
+
\frac{1}{|B|}
\sum_{b \in B}
\min_{a \in A} \|b - a\|_2^2.
\]
This avoids imposing an arbitrary order on scaffold, front, or boundary tokens.

\paragraph{Full objective.}
The full objective is
\[
\mathcal{L} =
\mathcal{L}_{\mathrm{pred}}
+ \lambda_{\mathrm{act}} \mathcal{L}_{\mathrm{act}}
+ \lambda_{\mathrm{energy}} \mathcal{L}_{\mathrm{energy}}
+ \lambda_{\mathrm{vc}} \mathcal{L}_{\mathrm{vc}}.
\]
We use $\lambda_{\mathrm{act}} = 0.05$, $\lambda_{\mathrm{energy}} = 0.02$, and $\lambda_{\mathrm{vc}} = 0.04$. $\mathcal{L}_{\mathrm{vc}}$ is a variance-covariance regularizer applied to global states and pooled token states to prevent representation collapse.

\paragraph{Task readouts.}
For global tasks, we pool
\[
\left\{
\hat{s}^{\mathrm{global}},
\mathrm{Pool}(\hat{S}^{\mathrm{mat}}),
\mathrm{Pool}(\hat{S}^{\mathrm{front}}),
\mathrm{Pool}(\hat{S}^{\mathrm{bdry}})
\right\}
\]
and apply an MLP head. This readout is used for topology classification, material regression, growth-stage recognition, and retrieval.

For dense correspondence, a token-level descriptor head predicts descriptors for mesh anchors. Descriptors are interpolated from anchors to vertices using barycentric weights, and correspondences are obtained by nearest-neighbor matching in descriptor space.

For mesh forecasting, we use a cross-attention geometry readout. Query tokens are formed from current mesh anchors plus $K_g = 256$ learned growth queries initialized near active front tokens. Each query attends to $\hat{S}_{t+\Delta t}$ and predicts a future position offset and validity score. Valid growth queries are locally triangulated with neighboring predicted anchors to reconstruct the forecast surface. This readout converts the predicted state into geometry for evaluation; the world model itself remains the state transition $S_t \mapsto \hat{S}_{t+\Delta t}$.

\paragraph{Common changing-support forecasting interface.}
Several comparison methods were originally designed for dynamics on fixed support. To compare their dynamics rather than penalize their native output parameterization, we retain each method's original dynamics backbone and attach the same variable-cardinality geometry interface used for forecasting comparisons. We report the results in Tab.~\ref{tab:capacity-matched-baselines}. Queries anchored to the current support predict updated positions for existing geometry, while $K_g=256$ learned growth queries predict candidate new vertices and validity scores. Valid candidates are reconstructed using the same local triangulation procedure.

Future vertex count, birth locations, and connectivity are withheld from every method. Each model is trained against the complete future surface and evaluated using the same Chamfer and vertex-drift protocols. Thus, the common readout standardizes how a latent prediction is converted into changing-support geometry while leaving the underlying state representation and dynamics model being compared intact.

\paragraph{Comparison protocol.}
Parent-level train/validation/test splits keep all counterfactual sibling branches from a parent trajectory in the same split. Unless an experiment explicitly varies an input or supervision source, comparison methods receive the same available observation modalities, material action and horizon inputs, training parents, augmentations, optimization epochs, and validation-trial budget. The matched allocation controls additionally equalize state-token count, parameter count, and transition computation.

\begin{table}[t]
\centering
\small
\begin{tabular}{lccccc}
\toprule
Method & Params & GPU-h & A$\uparrow$ & C$\downarrow$ & m$\uparrow$ \\
\midrule
ETNN-L      & 40M & 18.2 & 94.6 & --   & --   \\
MeshGPT-L   & 41M & 19.4 & --   & 42.9 & --   \\
Motion2Vec  & 40M & 18.7 & --   & 35.6 & --   \\
ULIP-2      & 41M & 18.9 & --   & --   & 49.1 \\
PointCLIP   & 40M & 18.1 & --   & --   & 50.3 \\
FOLIAGE     & 43M & 20.8 & \textbf{98.1} & \textbf{26.4} & \textbf{61.6} \\
\bottomrule
\end{tabular}
\caption{Capacity- and training-budget-matched task baselines. A is topology accuracy, C is 5-step direct forecasting Chamfer $\times10^3$, and m is retrieval mAP (\%). Scaling the comparison models does not close the task-specific gaps.}
\label{tab:capacity-matched-baselines}
\end{table}

These controls are cplementary to the matched state-allocation study in the main paper: the former tests whether larger task-specific baselines close the gap, while the latter isolates how a fixed FOLIAGE-sized state budget should be allocated and updated.

\paragraph{Inference path.}
At deployment, FOLIAGE uses only observable inputs. Given $x_t$, the context encoder produces fused tokens, GSA forms $S_t$, and the dynamics module predicts $\hat{S}_{t+\Delta t}$ under $a_t$ and $\Delta t$. The target encoder, physical energies, and energy heads are not used. If future observations become available, the state can be refreshed by re-encoding; otherwise, the model rolls out in latent space.

\paragraph{Complexity.}
Let $N = |U_t|$ be the number of fused input tokens and $E = |\mathcal{E}_t|$ the number of sparse correspondence edges. Token fusion costs $\mathcal{O}(E d)$ per sparse attention layer. GSA selects $K_m$ mature scaffold tokens, $K_f$ active front tokens, and $K_b$ interface tokens. Dynamics cost is dominated by front self-attention and front-interface cross-attention:
\[
\mathcal{O}\!\left(
K_f^2 d
+
K_b (K_f + K_m) d
+
K_m k d
\right),
\]
where $k$ is the scaffold graph neighborhood size. With default values $K_m = 128$, $K_f = 256$, and $K_b = 64$, the dynamics state is much smaller than the full mesh. GSA trades a modest amount of structured state for better allocation of modeling capacity, especially compared with full high-resolution token dynamics over all mesh anchors.

\section{Robustness Protocol Details}

We evaluated sensor robustness under four regimes. \textbf{Rich} uses RGB, point cloud, and mesh observations. \textbf{Typical} uses RGB and point cloud observations with incomplete mesh inputs. \textbf{Sparse} uses only one geometric modality. \textbf{Noisy} adds sensor noise and random modality dropout. The same downstream probe is used across all regimes so that performance differences reflect sensing degradation rather than task-specific tuning.

For correspondence corruption, \textbf{Corr-noise} rewires $20\%$ of pixel/point-to-mesh edges and drops $25\%$ of remaining cross-modal edges. \textbf{Occlusion} masks $50\%$ of RGB tokens and co-masks linked 3D tokens. \textbf{Drift} injects camera pose and intrinsics jitter before constructing cross-modal links. \textbf{Drift+SLAM} applies pose refinement before fusion. These settings test whether correspondence-aware fusion behaves as a robust grounding interface rather than an oracle dependency.

\section{Initialization Perturbation}

We additionally stress-test cross-modal retrieval under inaccurate initial mesh alignment. We perturb the initial mesh before constructing cross-modal links by applying translation proportional to the bounding-box diagonal and a random rotation. FOLIAGE degrades monotonically as perturbation increases, but remains ahead of the correspondence-free ablation and retrieval baselines under mild and medium perturbations. Pose refinement partially recovers performance in the strong-perturbation setting. The results are report in Tab.~\ref{tab:cross-model-retrieve-init-perturb}..

\begin{table}[ht]
\centering
\resizebox{0.95\textwidth}{!}{
\begin{tabular}{lcccc}
\toprule
\textbf{Init perturbation} & \textbf{FOLIAGE} & \textbf{w/o correspondence fusion} & \textbf{PointCLIPv2} & \textbf{ULIP-2} \\
\midrule
Clean                         & 0.616 & 0.502 & 0.480 & 0.460 \\
Mild: $5\%$ bbox + $3^\circ$  & $0.568 \pm 0.010$ & $0.486 \pm 0.009$ & $0.456 \pm 0.010$ & $0.444 \pm 0.009$ \\
Medium: $10\%$ bbox + $6^\circ$ & $0.511 \pm 0.012$ & $0.463 \pm 0.010$ & $0.414 \pm 0.011$ & $0.402 \pm 0.010$ \\
Strong: $20\%$ bbox + $12^\circ$ & $0.445 \pm 0.014$ & $0.420 \pm 0.012$ & $0.347 \pm 0.012$ & $0.338 \pm 0.011$ \\
Strong + pose refinement      & $0.485 \pm 0.013$ & $0.431 \pm 0.011$ & $0.361 \pm 0.012$ & $0.352 \pm 0.011$ \\
\bottomrule
\end{tabular}
}
\caption{Cross-modal retrieval under initialization perturbations.} \label{tab:cross-model-retrieve-init-perturb}
\end{table}

\section{Long-Horizon Rollout Stability}

We first evaluate recursive rollouts at horizons $(5, 10, 20, 40)$ in Tb.~\ref{tab:long-horizon-5-to-40}. FOLIAGE updates active growth-front tokens while preserving mature scaffold context, reducing drift accumulation over long horizons. Chamfer error increases with horizon for all methods, but FOLIAGE grows more slowly than the strongest baseline.

\begin{table}[ht]
\centering
\begin{tabular}{rcc}
\toprule
\textbf{Horizon} & \textbf{Best baseline Chamfer $\downarrow$} & \textbf{FOLIAGE Chamfer $\downarrow$} \\
\midrule
5  & 0.022 & \textbf{0.014} \\
10 & 0.029 & \textbf{0.022} \\
20 & 0.045 & \textbf{0.026} \\
40 & 0.075 & \textbf{0.041} \\
\bottomrule
\end{tabular}
\caption{Long-horizon rollout (recursive) stability measured by Chamfer distance.} \label{tab:long-horizon-5-to-40}
\end{table}

\paragraph{Full post-prefix trajectories.} \label{sec:full-rollout}
The 40-step experiment above targets the early regime in which recursive errors begin to compound. Each SURF-GARDEN trajectory, however, contains a 50-frame common prefix followed by 350 future frames. We therefore also evaluate the complete post-prefix trajectory by recursively applying the one-step transition without re-initializing the state from ground-truth future observations. The results are reported in Tab.~\ref{tab:full-trajectory}

\begin{table}[t]
\centering
\small
\begin{tabular}{lcccc}
\toprule
Method & C40$\downarrow$ & C100$\downarrow$ & C200$\downarrow$ & C350$\downarrow$ \\
\midrule
Motion2VecSets       & 75.4 & 158.2 & 293.6 & 489.1 \\
Uniform FPS-448 AR   & 62.7 & 133.2 & 250.9 & 415.1 \\
FOLIAGE w/o GSA     & 52.7 & 103.1 & 190.2 & 319.6 \\
\textbf{FOLIAGE}     & \textbf{41.2} & \textbf{82.4} & \textbf{153.7} & \textbf{261.8} \\
\bottomrule
\end{tabular}
\caption{Full post-prefix recursive rollout. Values are Chamfer $\times10^3$. Absolute error grows for every model, but FOLIAGE accumulates error more slowly over the complete 350-frame future.}
\label{tab:full-trajectory}
\end{table}

All models accumulate substantial error over sufficiently long open-loop prediction, as expected. The ordering nevertheless remains stable throughout the trajectory: growth-aware allocation reduces error relative to homogeneous FOLIAGE, and the full scaffold/front/interface dynamics further reduce accumulation. The advantage therefore does not arise only from short-horizon prediction.

\section{Distributed Growth and Activity Localization}
\label{sec:distributed-growth}

The term ``growth front'' denotes a functional state role rather than a geometric boundary. A future-active region may occur anywhere on the surface and may consist of multiple disconnected or interior regions. To characterize the regime in which adaptive allocation is useful, we vary the fraction of the surface undergoing substantial future change while holding topology, material distribution, sensing, and state-token budget fixed.

\begin{table}[t]
\centering
\small
\begin{tabular}{lccc}
\toprule
Method &
Localized 10\% C40$\downarrow$ &
Diffuse 50\% C40$\downarrow$ &
Global 100\% C40$\downarrow$ \\
\midrule
Uniform FPS &
$62.7\pm1.8$ &
$76.9\pm2.1$ &
$107.8\pm3.1$ \\
FOLIAGE w/o GSA &
$52.7\pm1.4$ &
$72.1\pm2.0$ &
$106.3\pm3.0$ \\
\textbf{FOLIAGE} &
$\mathbf{41.2\pm1.1}$ &
$\mathbf{61.9\pm1.7}$ &
$\mathbf{103.6\pm2.9}$ \\
\midrule
Activity Recall@256 &
0.924 &
0.712 &
0.512 \\
\bottomrule
\end{tabular}
\caption{Effect of increasingly distributed growth. C40 is recursive Chamfer $\times10^3$; mean $\pm$ standard deviation is over five independently trained seeds. Activity Recall@256 measures coverage of future-active surface by the 256 selected front tokens.}
\label{tab:distributed-growth}
\end{table}

In Tab.~\ref{tab:distributed-growth}, the absolute forecasting error rises for every model as a larger fraction of the surface changes. FOLIAGE's advantage is largest under spatially nonuniform growth, where a bounded state budget can concentrate on a comparatively small future-active region. As activity becomes global, activity recall necessarily decreases and uniform allocation becomes increasingly appropriate, causing the gap to narrow. Error analysis shows that the largest local errors occur in high-change regions omitted by the activity selector, identifying activity mislocalization as a principal failure mode of GSA.

\section{Transfer Protocol Details}

For GrowliFlower, we evaluate growth-stage recognition using a frozen FOLIAGE encoder, a few-shot linear probe, and light fine-tuning of the last encoder block. For Pheno4D and Crops3D, correspondences are initialized using anchored ICP with pruning and temporal pose smoothing before applying learned descriptors. For D-FAUST and CAPE, we use the standard correspondence structure provided by the datasets and evaluate whether the learned descriptors transfer to dynamic 3D surfaces beyond the growing-surface simulator.

These protocols test external validity without relying on simulator oracle correspondences. The controlled simulator provides counterfactual supervision and hidden physical quantities; transfer experiments test whether the learned representation remains useful when those simulator-specific signals are absent.

\begin{figure*}[ht]
    \centering
    \setlength\fboxrule{2pt}
    \setlength{\tabcolsep}{6pt}
    \renewcommand{\arraystretch}{1.2}
    \begin{tabular}{ccc}
        \tikz[baseline]{\node[inner sep=0] 
            {\includegraphics[width=0.25\textwidth]{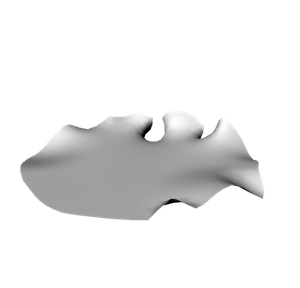}};} &
        \tikz[baseline]{\node[draw=black, thick, inner sep=0]
            {\includegraphics[width=0.25\textwidth]{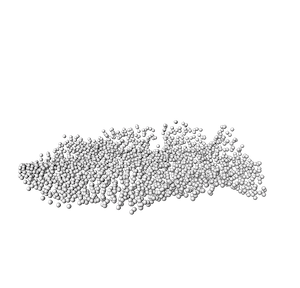}};} &
        \tikz[baseline]{\node[draw=black, dashed, thick, inner sep=0] 
            {\includegraphics[width=0.25\textwidth]{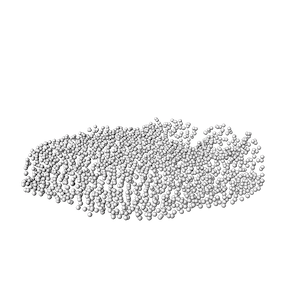}};} \\
        \small Ground Truth & \small 1 & \small 2 \\
        \tikz[baseline]{\node[draw=black, dashed, thick, inner sep=0]
            {\includegraphics[width=0.25\textwidth]{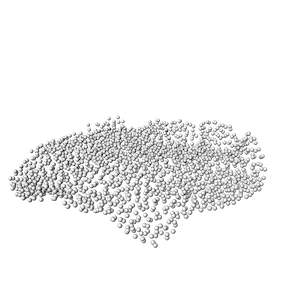}};} &
        \tikz[baseline]{\node[draw=black, dashed, thick, inner sep=0]
            {\includegraphics[width=0.25\textwidth]{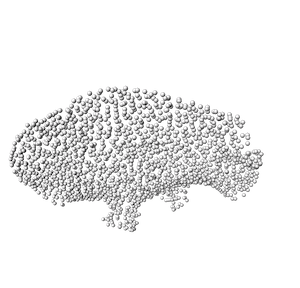}};} &
        \tikz[baseline]{\node[draw=black, dashed, thick, inner sep=0]
            {\includegraphics[width=0.25\textwidth]{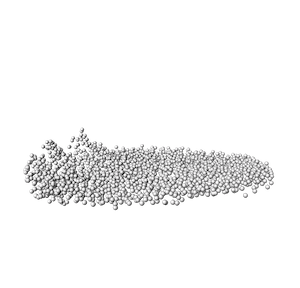}};} \\
        \small 3 & \small 4 & \small 5 \\
    \end{tabular}
    \caption{Top-5 retrievals on SURF-BENCH (Image $\rightarrow$ Point Cloud)} \label{fig:t5-1}
\end{figure*} 

\paragraph{Passive real-world forecasting protocol.}
We use the same baselines and report the resulting forecast errors in Tab.~\ref{tab:t6-real-forecast}. Chamfer distance is normalized by the plant bounding box and reported as $\times10^3$. Real forecasting results average three fine-tuning seeds; confidence intervals are 95\% bootstrap intervals over held-out plant trajectories.

The two-week cabbage setting contains large changes in plant size, leaf pose, and newly observed surface. FOLIAGE obtains newly observed-surface F1 of 55.3\% [53.1,57.5], compared with 28.7\% [26.6,30.8] for persistence. This test therefore requires more than retaining the geometry already visible at the input time. At the same time, these datasets provide only naturally occurring passive growth: SURF-GARDEN remains complementary because it provides matched initial states under alternative material interventions.

\begin{table*}[t]
\centering
\begin{tabular}{@{}lclclc@{}}
\toprule
\textbf{Latent Dim. \(d\)} & \textbf{Score} &
\textbf{EMA Rate} & \textbf{Score} &
\textbf{Sampling Range (\(\Delta t\))} & \textbf{Score} \\
\midrule
512 & 0.74 &
0.995 & 0.80 &
Uniform 1–4 & 0.78 \\
640 & 0.79 &
0.997 & 0.81 &
Uniform 1–12 & 0.80 \\
768 (Ours) & \textbf{0.82} &
0.998 (Ours) & \textbf{0.82} &
Uniform 1–20 (Ours) & \textbf{0.82} \\
896 & 0.81 &
0.999 & 0.81 &
Uniform 1–28 & 0.81 \\
1024 & 0.78 &
0.9995 & 0.79 &
Uniform 1–36 & 0.79 \\
\bottomrule
\end{tabular}
\vspace{0.5em}
\caption{Ablation results for model capacity and temporal encoding. Each block shows the effect of sweeping a single hyperparameter on the composite validation score.}
\label{tab:model-capacity-encoding}
\end{table*}

\begin{table*}[t]
\centering
\small
\begin{tabular}{lccc}
\toprule
Method &
Material MAE $\times10^3$ $\downarrow$ &
Direct C5 $\times10^3$ $\downarrow$ &
Recursive C40 $\times10^3$ $\downarrow$ \\
\midrule
DiffCloth &
$53.1\pm1.1$ [52.2,54.0] &
-- &
-- \\
Motion2VecSets &
-- &
$38.1\pm0.7$ [37.5,38.7] &
$75.4\pm2.2$ [73.5,77.3] \\
Uniform FPS-448 &
$35.2\pm0.8$ [34.6,35.9] &
$28.2\pm0.5$ [27.8,28.7] &
$62.7\pm1.8$ [61.2,64.3] \\
FOLIAGE w/o GSA &
$35.0\pm0.7$ [34.4,35.6] &
$28.0\pm0.5$ [27.6,28.4] &
$52.7\pm1.4$ [51.5,54.0] \\
\textbf{FOLIAGE} &
$\mathbf{32.1\pm0.6}$ [31.6,32.7] &
$\mathbf{25.7\pm0.4}$ [25.3,26.1] &
$\mathbf{41.2\pm1.1}$ [40.3,42.2] \\
\bottomrule
\end{tabular}
\caption{Statistical reliability of the main hidden-material and forecasting results. C5 is direct prediction at five steps; C40 is a recursive 40-step rollout.}
\label{tab:statistical-reliability}
\end{table*}

\section{SURF-GARDEN Details}

In SURF-GARDEN, surfaces undergo \emph{metric accretion}, where new material is added, and deform like thin elastic shells. Energy-based simulations produce FEM-quality meshes with two sensor projections per frame. With rich physics, precise cross-modal alignment, and topological diversity, SURF-GARDEN offers a well-rounded addition to accretive surface growth research.

\subsection{SURF-GARDEN} \label{sec:dataset-sim}

\textbf{Energy Model.} Formally, SURF-GARDEN embeds a non-Euclidean metric into $\mathbb{R}^3$, inducing negative curvature and curling, an approach grounded in differential geometry and plant morphogenesis research. For a mesh $\mathcal{M}_t = (V_t, E_t, F_t)$, the simulator minimizes a smooth energy \cite{grinspun2003discrete}:

\vspace{-1.2em}
\[
\mathcal{E}(V_t) = 
\underbrace{k_\text{stretch} \textstyle \sum\nolimits_{e} (\|e\| - \ell_e^{\star})^2 +
k_\text{shear} \textstyle \sum\nolimits_{f} \|S_f\|_F^2}_\mathcal{E_\text{membrane}} +
\underbrace{k_\text{bend} \textstyle \sum\nolimits_{(f_i, f_j)} (\theta_{ij} - \theta_{ij}^\star)^2}_\mathcal{E_\text{flexural}},
\]
\vspace{-1.0em}

where $\ell_e^\star$ is the rest length, $S_f$ the shear tensor, and $\theta_{ij}^\star = 0$ the preferred dihedral. The closed-form gradients \cite{tamstorf2013discrete} are efficient to evaluate and support stable forward Euler steps at $\Delta t = 10^{-2}$.

\textbf{Metric Accretion and Non-Euclidean Growth.} Growth is modulated by $g(v) \in [0,1]$, the normalized geodesic distance to a source set, simulating hormone diffusion (e.g., auxin). Rest lengths update as $\ell_e^\star(t) = ( (g(v_i) + g(v_j))/{2} + 1 )\|e\|$. Edges that exceed $1.5\times$ their original rest length are split. This guided adaptive refinement varies growth spatially in a differential manner, with some parts growing faster than others. Coefficients $k_\text{stretch}, k_\text{shear}, k_\text{bend}$ introduce temporal variation.

\textbf{Mesh Quality.} Meshes are optimized with ODT smoothing \cite{chen2004mesh} and Delaunay edge flips, producing near-isotropic triangles ($0.88 \pm 0.04$ radius-edge ratio) and valence $5\!-\!7$. Self-intersections are avoided using ellipsoidal vertex colliders in a bounding-volume hierarchy (BVH) \cite{wald2007fast}.

\textbf{Topological Variety.} We include six classes of 2-manifolds with boundary: disc, annulus, punctured torus, Möbius strip, pair-of-pants, and thrice-punctured disc. These cover lobed, twisted, and compound forms common in botany and geometry alike, reducing overfitting to select shapes.

\textbf{Counterfactual Branching.} Each sequence begins with a 50-frame prefix under baseline parameters $\theta^A$, after which it branches into two or more trajectories with modified controls $\theta^B, \theta^C, \dots$ (e.g., halved $k_\text{bend}$). Identifiers in the half-edge structure ensure that vertex indices and geometry remain aligned post-branch. This enables supervised counterfactual reasoning: identical pasts yield distinct futures, and the model must predict each outcome conditioned on an action token.

Such branch-point supervision equips physically intelligent agents with the ability to forecast the consequences of their own interventions. During training, FOLIAGE observes the prefix through the context encoder and rolls out to each branch using its corresponding action tokens. At inference, alternate futures can be queried by swapping tokens—no simulator calls are required.

SURF-GARDEN provides $7,200$ branched growth sequences, each defined by $k_\text{stretch}, k_\text{shear}, k_\text{bend}$, a topology class, and a random seed. Every sequence spans $400$ frames, evolving from rest to maturity with vertex counts that increase from 20 to $10^5$. Variation is further introduced through quaternion perturbations and vector fields. We split the dataset $8\!:\!1\!:\!1$ into train/val/test.

\subsection{Dataset correspondence and annotation} \label{sec:dataset-correspond}

Each frame yields two mesh-tied modalities. \textit{(1) Multi-view RGB}: eight cameras on a Fibonacci sphere render photorealistic frames with Cycles shading \cite{blender2023}, HDR lighting, and 50mm lenses. Exposure jitter, defocus, and 20\% CutOut masking improve robustness \cite{devries2017improved}. Each pixel carries its emitting triangle index and barycentric coordinates. Cameras are fixed per trajectory; masks persist over time. \textit{(2) LiDAR-style point cloud}: a $64 \times 2048$ raycast with $\sigma{=}5$\,mm Gaussian noise and 5\% dropout mimics real-world sparsity. Each point stores its nearest mesh vertex. Both views preserve consistent token indices, enabling direct cross-modal supervision.

Across frames, a half-edge data structure is maintained with a unique identifier for the vertices, edges, and faces. Even as vertices and edges are added (new vertices, edge flips etc.) and their indices are updated, these identifiers remain the same, allowing us to exactly identify the same mesh elements over time alongside their quantities of interest (energy, age, etc.).

\begin{figure*}[ht]
    \centering
    \setlength\fboxrule{2pt}
    \setlength{\tabcolsep}{6pt}
    \renewcommand{\arraystretch}{1.2}
    \begin{tabular}{ccc}
        \tikz[baseline]{\node[inner sep=0] 
            {\includegraphics[width=0.25\textwidth]{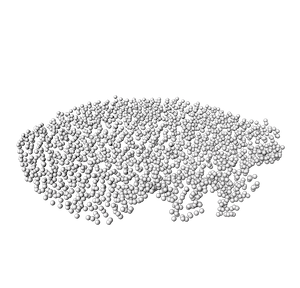}};} &
        \tikz[baseline]{\node[draw=black, thick, inner sep=0]
            {\includegraphics[width=0.25\textwidth]{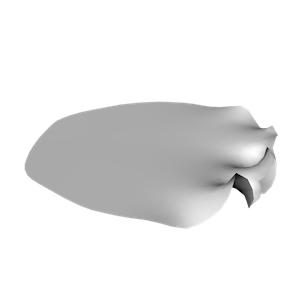}};} &
        \tikz[baseline]{\node[draw=black, dashed, thick, inner sep=0] 
            {\includegraphics[width=0.25\textwidth]{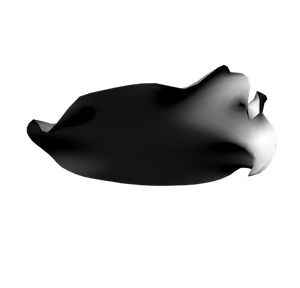}};} \\
        \small Ground Truth & \small 1 & \small 2 \\
        \tikz[baseline]{\node[draw=black, dashed, thick, inner sep=0]
            {\includegraphics[width=0.25\textwidth]{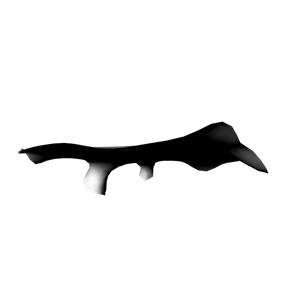}};} &
        \tikz[baseline]{\node[draw=black, dashed, thick, inner sep=0]
            {\includegraphics[width=0.25\textwidth]{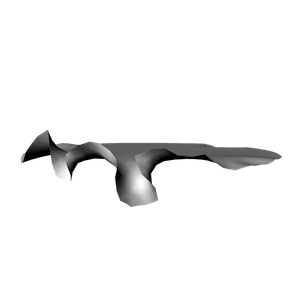}};} &
        \tikz[baseline]{\node[draw=black, dashed, thick, inner sep=0]
            {\includegraphics[width=0.25\textwidth]{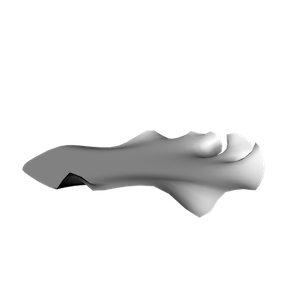}};} \\
        \small 3 & \small 4 & \small 5 \\
    \end{tabular}
    \caption{Top-5 retrievals on SURF-BENCH (Point Cloud $\rightarrow$ Image)} \label{fig:t5-2}
\end{figure*}

\section{Critic Heads}
For each SURF-BENCH task, we attach a specialized critic to evaluate the core objective under realistic sensor constraints. The topology critic ingests mesh-only latent state into a frozen backbone plus a 768$\rightarrow$256$\rightarrow$6 MLP for genus classification; the material critic uses a single RGB-based latent with a one-hidden-layer regressor to predict bending modulus; the stage critic processes four latent state embeddings through a 128-unit Bi-GRU \cite{cho2014learning} for balanced growth-stage accuracy; the growth critic conditions MeshGPT's \cite{siddiqui2024meshgpt} autoregressive split/offset tokens on $M_t$ and $s_{t + \Delta t}$ to measure Chamfer \cite{barrow1977parametric} and vertex-count errors; the retrieval critic ranks image-to-mesh latent states by cosine similarity \cite{salton1983introduction}; and the correspondence critic refines per-vertex features with global and young-region tokens via a 2-layer residual MLP, projects onto 128 spectral components, solves a functional map \cite{ovsjanikov2012functional}, and applies five ZoomOut refinements \cite{melzi2019zoomout}.

\section{Action Encoder}
\label{sec:model-action}

The action space in our unbounded surface evolution settings consists of the three scalar elastic coefficients that
parameterise the shell mechanics,
\(\mathbf{a}_t^{\mathrm{raw}}=[k_{\mathrm{stretch}},\,k_{\mathrm{shear}},\,k_{\mathrm{bend}}]\).
Because these values span several orders of magnitude, we first apply a
logarithmic re–scaling $\tilde{k}=\log_{10}(k), \, k\in\{k_{\mathrm{stretch}},k_{\mathrm{shear}},k_{\mathrm{bend}}\}$ followed by z–score normalization using the mean and variance estimated throughout the training set.  The normalized vector
\(\tilde{\mathbf a}_t\in\mathbb R^{3}\) is then embedded into the model’s token
space through a two–layer perceptron $\mathbf a_t = \operatorname{MLP}_{\mathrm{act}}(\tilde{\mathbf a}_t), \operatorname{MLP}_{\mathrm{act}}:\;3 \rightarrow 128 \rightarrow d,\;d=768$ with GELU activations \cite{hendrycks2016gaussian} and layer normalization \cite{ba2016layer}. This produces the action
token \(\mathbf a_t\in\mathbb R^{768}\).

At training time, the encoder also encounters a learned $\mathbf a_\text{null}$ embedding that substitutes for $\mathbf a_t$ whenever
material coefficients are withheld. During training we drop the entire action token with probability $0.1$ for this scenario. During inference, the user may supply a physical–coefficient vector to perform counterfactual
roll–outs; if omitted, the encoder inserts the null token, reverting the model
to passive prediction behavior.

\section{Mesh Forecasting}

Fig.~\ref{fig:t4-1}, \ref{fig:t4-2}, and \ref{fig:t4-3} shows examples of mesh prediction by FOLIAGE and baselines across different look-ahead and action conditioning. For simple topology and limited growth, the general morphology of the surface is preserved. But near the boundaries and in areas of high feature activity (emergence or disappearing of buckling), prediction error increases, especially for baseline models. In Fig.~\ref{fig:t4-2}, baselines such as MeshGraphNets \cite{Pfaff2021_MeshGraphNets} and CaDeX \cite{Lei2022CaDeX} struggle to model surfaces that had enlarged substantially through accretive growth, resulting in visibly 'shrunk-down' predictions. With more complex topology such as the Möbius strip (Fig.~\ref{fig:t4-3}), these errors propagate globally. These disparities highlight the effectiveness of FOLIAGE's perception-action setup and physics-guided learning to model the complex deformations and growth of the surfaces.

\section{Cross-Model Retrieval}

\begin{figure*}[ht]
    \centering
    \setlength\fboxrule{2pt}
    \setlength{\tabcolsep}{6pt}
    \renewcommand{\arraystretch}{1.2}
    \begin{tabular}{ccc}
        \tikz[baseline]{\node[inner sep=0] 
            {\includegraphics[width=0.25\textwidth]{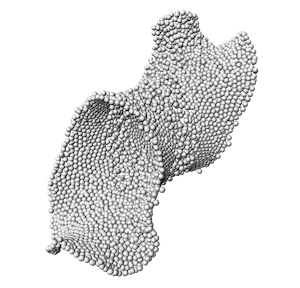}};} &
        \tikz[baseline]{\node[draw=black, thick, inner sep=0]
            {\includegraphics[width=0.25\textwidth]{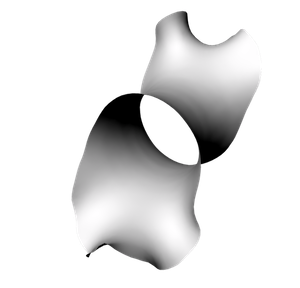}};} &
        \tikz[baseline]{\node[draw=black, dashed, thick, inner sep=0] 
            {\includegraphics[width=0.25\textwidth]{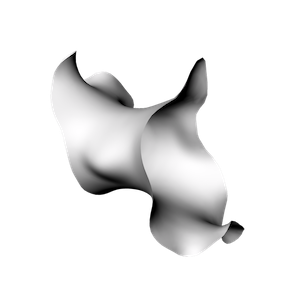}};} \\
        \small Ground Truth & \small 1 & \small 2 \\
        \tikz[baseline]{\node[draw=black, dashed, thick, inner sep=0]
            {\includegraphics[width=0.25\textwidth]{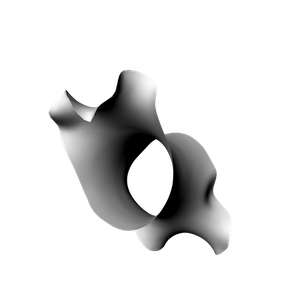}};} &
        \tikz[baseline]{\node[draw=black, dashed, thick, inner sep=0]
            {\includegraphics[width=0.25\textwidth]{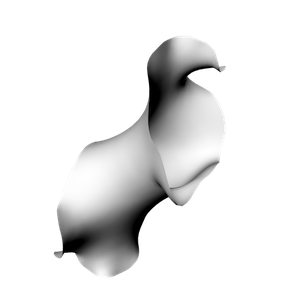}};} &
        \tikz[baseline]{\node[draw=black, dashed, thick, inner sep=0]
            {\includegraphics[width=0.25\textwidth]{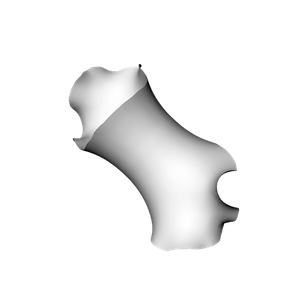}};} \\
        \small 3 & \small 4 & \small 5 \\
    \end{tabular}
    \caption{Top-5 retrievals on SURF-BENCH (Point Cloud $\rightarrow$ Image, Zero-shot)} \label{fig:t5-3}
\end{figure*} 

\begin{figure*}[ht]
    \centering
    \setlength\fboxrule{2pt}
    \setlength{\tabcolsep}{6pt}
    \renewcommand{\arraystretch}{1.2}
    \begin{tabular}{ccc}
        \tikz[baseline]{\node[inner sep=0] 
            {\includegraphics[width=0.25\textwidth]{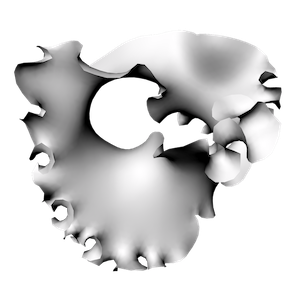}};} &
        \tikz[baseline]{\node[draw=black, thick, inner sep=0]
            {\includegraphics[width=0.25\textwidth]{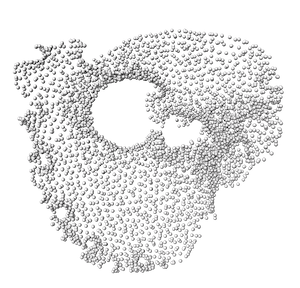}};} &
        \tikz[baseline]{\node[draw=black, dashed, thick, inner sep=0] 
            {\includegraphics[width=0.25\textwidth]{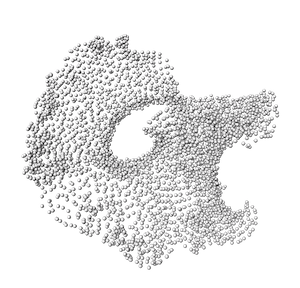}};} \\
        \small Ground Truth & \small 1 & \small 2 \\
        \tikz[baseline]{\node[draw=black, dashed, thick, inner sep=0]
            {\includegraphics[width=0.25\textwidth]{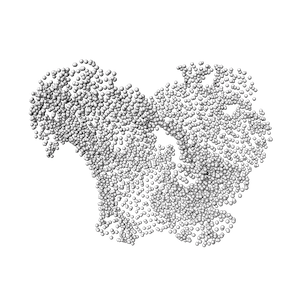}};} &
        \tikz[baseline]{\node[draw=black, dashed, thick, inner sep=0]
            {\includegraphics[width=0.25\textwidth]{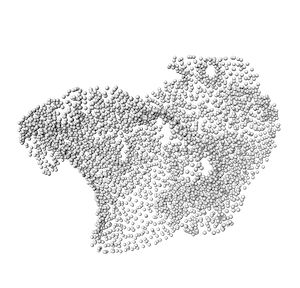}};} &
        \tikz[baseline]{\node[draw=black, dashed, thick, inner sep=0]
            {\includegraphics[width=0.25\textwidth]{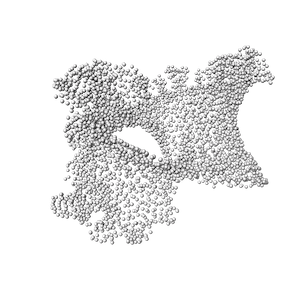}};} \\
        \small 3 & \small 4 & \small 5 \\
    \end{tabular}
    \caption{Top-5 retrievals on SURF-BENCH (Image $\rightarrow$ Point Cloud, Zero-shot)} \label{fig:t5-4}
\end{figure*} 

In Fig.~\ref{fig:t5-1}, \ref{fig:t5-2}, \ref{fig:t5-3}, and \ref{fig:t5-4} we show examples of cross-model retrieval in normal and zero-shot settings for point clouds and images. The correct option (solid line border) is differentiated from the incorrect ones (dashed line border). We note that the purple lines highlighting the boundary (e.g. in Fig.~\ref{fig:t4-1}) are a visual aid; they are not present in the rendered images of the mesh surface. FOLIAGE's first choice is predominantly the correct one followed by visually similar surfaces (image rendering or point cloud representations) with the same topology categorization. This is observed for unseen examples with complex morphology and challenging viewing angles, indicating the strong semantic awareness and consistency of FOLIAGE's latent state across different modalities.

\section{Dense Correspondence}

\begin{figure*}[ht]
    \centering
    \setlength{\tabcolsep}{1pt}
    \begin{minipage}[t]{0.3\textwidth}
        \centering
        \includegraphics[width=\textwidth]{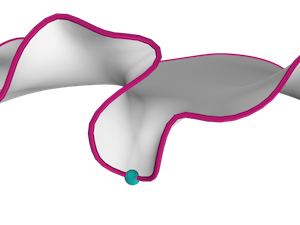}\\
        \small mesh at $t=100$
    \end{minipage}
    \hfill
    \begin{minipage}[t]{0.3\textwidth}
        \centering
        \includegraphics[width=\textwidth]{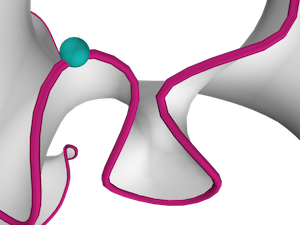}\\
        \small (b) mesh at $t=108$ (FOLIAGE) \label{fig:latent-rollout}
    \end{minipage}
    \caption{A vertex feature (turquoise sphere) that began at the bottom of a valley (left) quickly develops into the peak of a mountain (right).}
    \label{fig:t6-lm}
\end{figure*}

\begin{figure*}[t]
\centering
\setlength{\tabcolsep}{1pt}
\renewcommand{\arraystretch}{1.2}
\begin{tabular}{cccc}
    \includegraphics[width=0.24\textwidth]{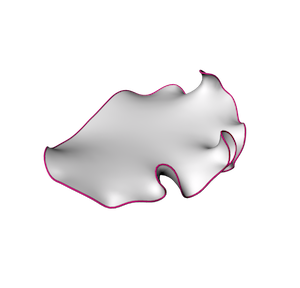} &
    \includegraphics[width=0.24\textwidth]{figs-supp/t6-1-s-a.png} &
    \includegraphics[width=0.24\textwidth]{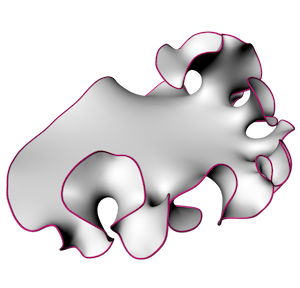} &
    \includegraphics[width=0.24\textwidth]{figs-supp/t6-1-t-a.png} \\
    \multicolumn{2}{c}{\small Source view A} & \multicolumn{2}{c}{\small Target (GT) view A} 
\end{tabular}

\vspace{2em}

\begin{tabular}{ccccc}
    \includegraphics[width=0.18\textwidth]{figs-supp/t6-1-ours-a.png} &
    \includegraphics[width=0.18\textwidth]{figs-supp/t6-1-b1-a.png} &
    \includegraphics[width=0.18\textwidth]{figs-supp/t6-1-b2-a.png} &
    \includegraphics[width=0.18\textwidth]{figs-supp/t6-1-b3-a.png} &
    \includegraphics[width=0.18\textwidth]{figs-supp/t6-1-b4-a.png} \\
    \small Ours & \small \cite{cao2024spectral} & \small DiffusionNet & \small G-MSM  & \small ZoomOut  \\
\end{tabular}

\vspace{2em}

\begin{tabular}{cccc}
    \includegraphics[width=0.24\textwidth]{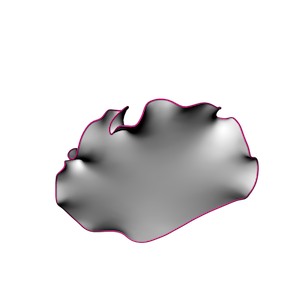} &
    \includegraphics[width=0.24\textwidth]{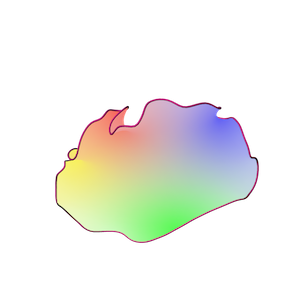} &
    \includegraphics[width=0.24\textwidth]{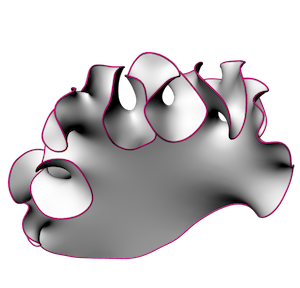} &
    \includegraphics[width=0.24\textwidth]{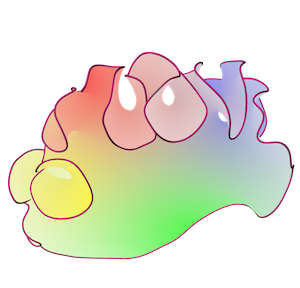} \\
    \multicolumn{2}{c}{\small Source view B} & \multicolumn{2}{c}{\small Target (GT) view B} 
\end{tabular}

\vspace{2em}

\begin{tabular}{ccccc}
    \includegraphics[width=0.18\textwidth]{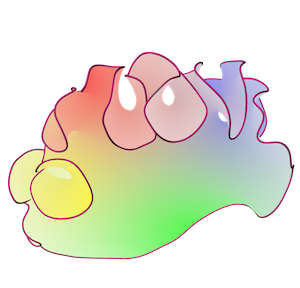} &
    \includegraphics[width=0.18\textwidth]{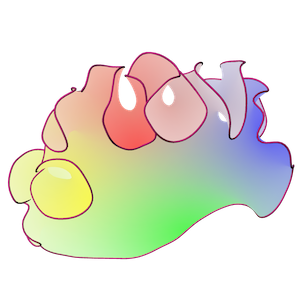} &
    \includegraphics[width=0.18\textwidth]{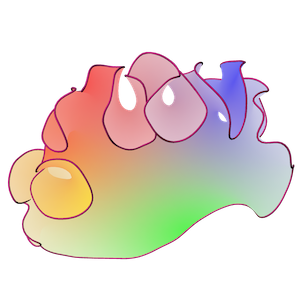} &
    \includegraphics[width=0.18\textwidth]{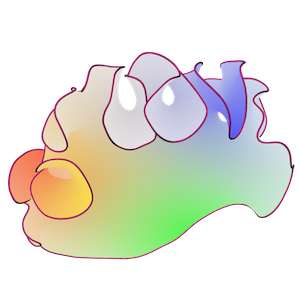} &
    \includegraphics[width=0.18\textwidth]{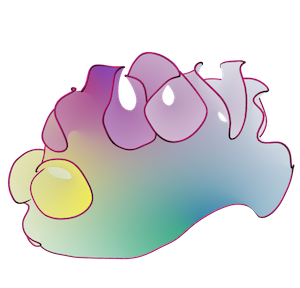} \\
    \small Ours & \small \cite{cao2024spectral} & \small DiffusionNet & \small G-MSM  & \small ZoomOut \\
\end{tabular}

\caption{Correspondences on SURF-BENCH.} \label{fig:t6-1}
\end{figure*} 

\begin{figure*}[t]
\centering
\setlength{\tabcolsep}{1pt}
\renewcommand{\arraystretch}{1.2}
\begin{tabular}{cccc}
    \includegraphics[width=0.24\textwidth]{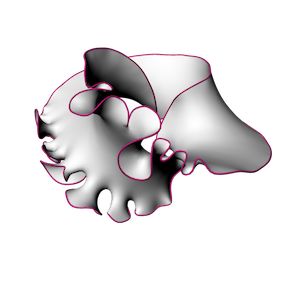} &
    \includegraphics[width=0.24\textwidth]{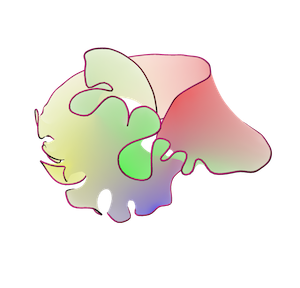} &
    \includegraphics[width=0.24\textwidth]{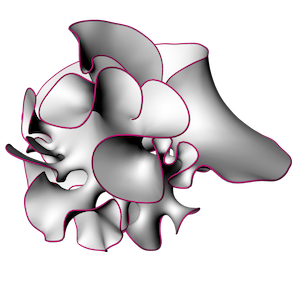} &
    \includegraphics[width=0.24\textwidth]{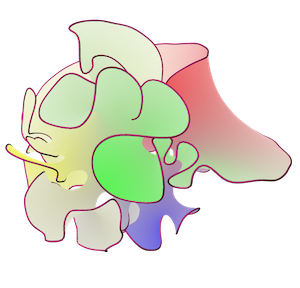} \\
    \multicolumn{2}{c}{\small Source view A} & \multicolumn{2}{c}{\small Target (GT) view A} 
\end{tabular}

\vspace{2em}

\begin{tabular}{ccccc}
    \includegraphics[width=0.18\textwidth]{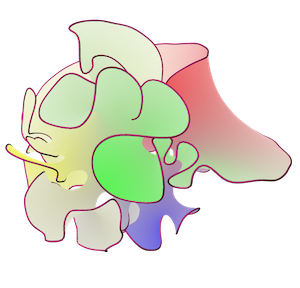} &
    \includegraphics[width=0.18\textwidth]{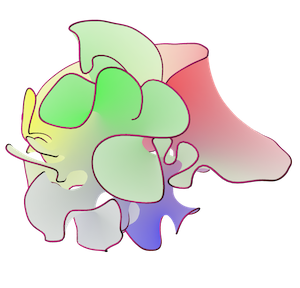} &
    \includegraphics[width=0.18\textwidth]{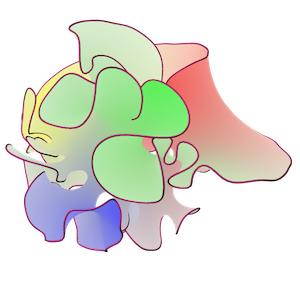} &
    \includegraphics[width=0.18\textwidth]{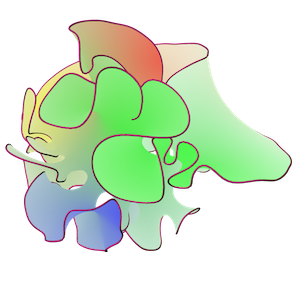} &
    \includegraphics[width=0.18\textwidth]{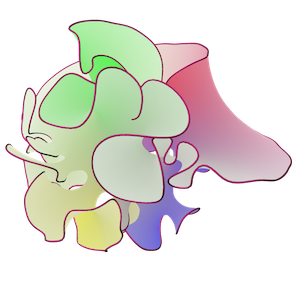} \\
    \small Ours & \small \cite{cao2024spectral} & \small DiffusionNet & \small G-MSM & \small ZoomOut\\
\end{tabular}

\vspace{2em}

\begin{tabular}{cccc}
    \includegraphics[width=0.24\textwidth]{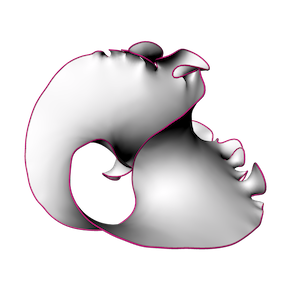} &
    \includegraphics[width=0.24\textwidth]{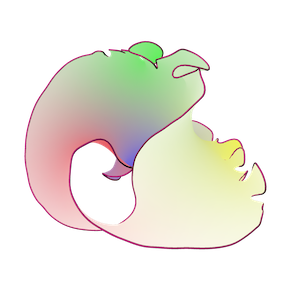} &
    \includegraphics[width=0.24\textwidth]{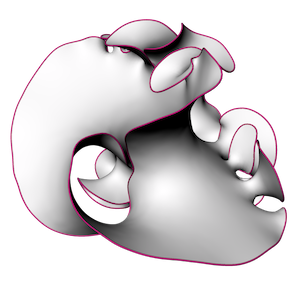} &
    \includegraphics[width=0.24\textwidth]{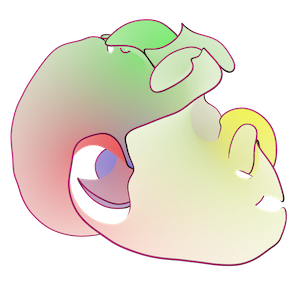} \\
    \multicolumn{2}{c}{\small Source view B} & \multicolumn{2}{c}{\small Target (GT) view B} 
\end{tabular}

\vspace{2em}

\begin{tabular}{ccccc}
    \includegraphics[width=0.18\textwidth]{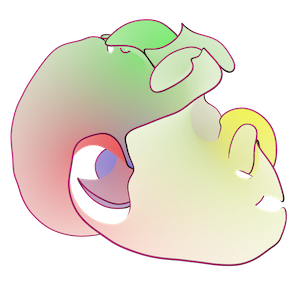} &
    \includegraphics[width=0.18\textwidth]{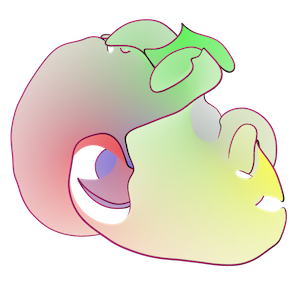} &
    \includegraphics[width=0.18\textwidth]{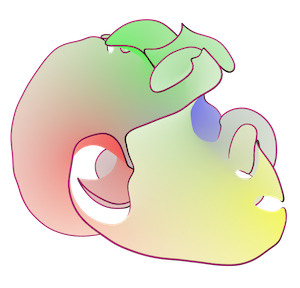} &
    \includegraphics[width=0.18\textwidth]{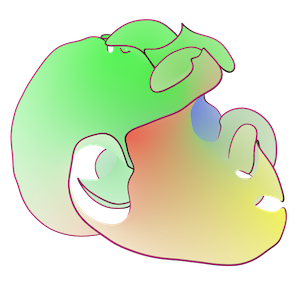} &
    \includegraphics[width=0.18\textwidth]{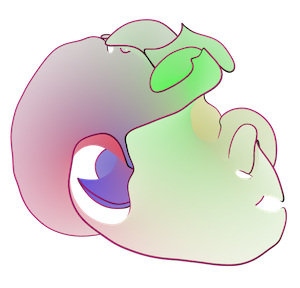} \\
    \small Ours & \small \cite{cao2024spectral} & \small DiffusionNet & \small G-MSM  & \small ZoomOut \\
\end{tabular}

\caption{Correspondences on SURF-BENCH.} \label{fig:t6-2}
\end{figure*}

\begin{figure*}[t]
\centering
\setlength{\tabcolsep}{1pt}
\renewcommand{\arraystretch}{1.2}
\begin{tabular}{cccc}
    \includegraphics[width=0.24\textwidth]{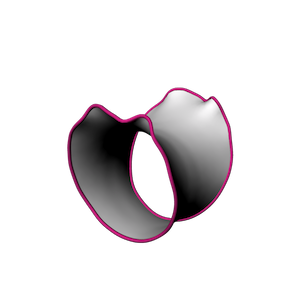} &
    \includegraphics[width=0.24\textwidth]{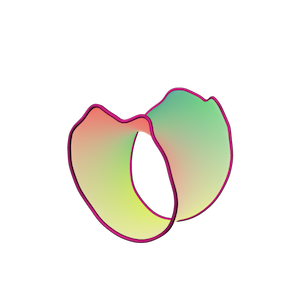} &
    \includegraphics[width=0.24\textwidth]{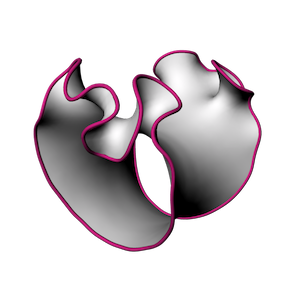} &
    \includegraphics[width=0.24\textwidth]{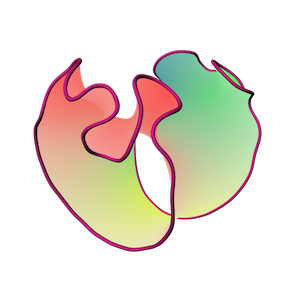} \\
    \multicolumn{2}{c}{\small Source view A} & \multicolumn{2}{c}{\small Target (GT) view A} 
\end{tabular}

\vspace{2em}

\begin{tabular}{ccccc}
    \includegraphics[width=0.18\textwidth]{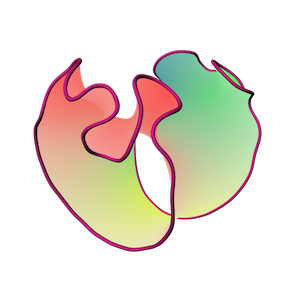} &
    \includegraphics[width=0.18\textwidth]{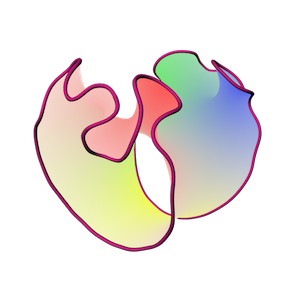} &
    \includegraphics[width=0.18\textwidth]{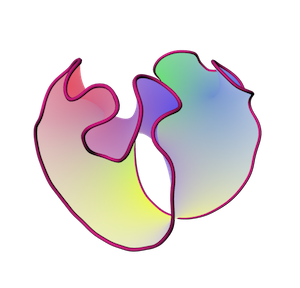} &
    \includegraphics[width=0.18\textwidth]{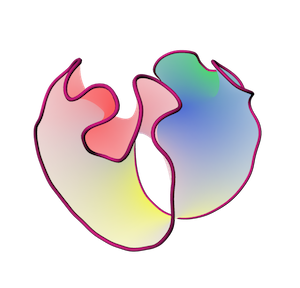} &
    \includegraphics[width=0.18\textwidth]{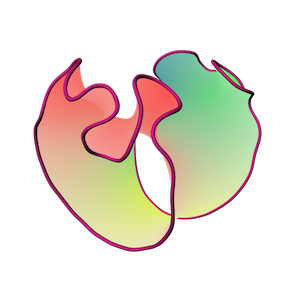} \\
    \small Ours & \small \cite{cao2024spectral} & \small DiffusionNet & \small G-MSM & \small ZoomOut \\
\end{tabular}

\vspace{2em}

\begin{tabular}{cccc}
    \includegraphics[width=0.24\textwidth]{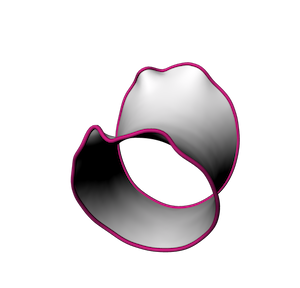} &
    \includegraphics[width=0.24\textwidth]{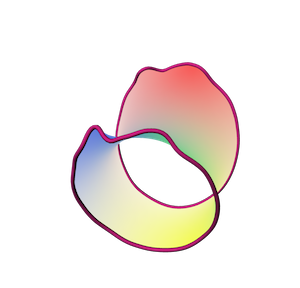} &
    \includegraphics[width=0.24\textwidth]{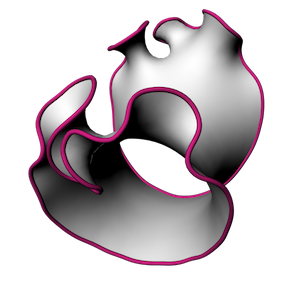} &
    \includegraphics[width=0.24\textwidth]{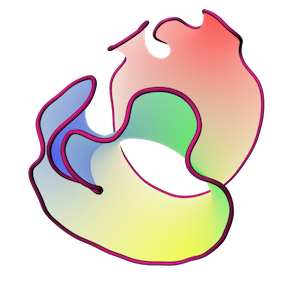} \\
    \multicolumn{2}{c}{\small Source view B} & \multicolumn{2}{c}{\small Target (GT) view B} 
\end{tabular}

\vspace{2em}

\begin{tabular}{ccccc}
    \includegraphics[width=0.18\textwidth]{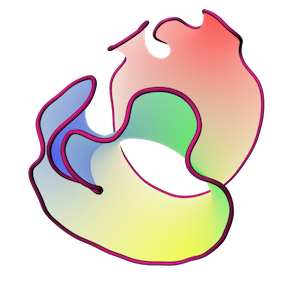} &
    \includegraphics[width=0.18\textwidth]{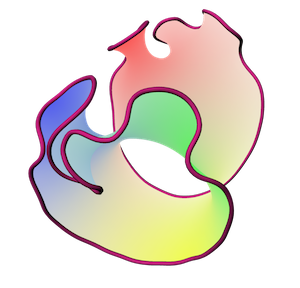} &
    \includegraphics[width=0.18\textwidth]{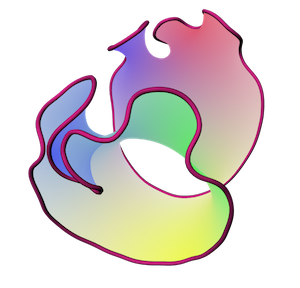} &
    \includegraphics[width=0.18\textwidth]{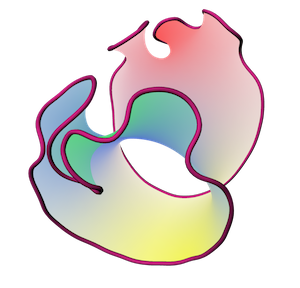} &
    \includegraphics[width=0.18\textwidth]{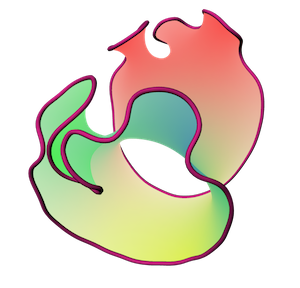} \\
    \small Ours & \small \cite{cao2024spectral} & \small DiffusionNet & \small G-MSM & \small ZoomOut \\
\end{tabular}

\caption{Correspondences on SURF-BENCH.} \label{fig:t6-3}
\end{figure*}

In contrast to well-studied domains like human or animal bodies, which follow a set template (the skeleton) movements constrained to specific parts of the geometry (e.g. arm movements has very limited impact on the full body), SURF-GARDEN's open surfaces represent a continuum in which features smoothly emerge and dissipate. As Fig.~\ref{fig:t6-lm} suggests, identifying one of a fixed number of extrusions (e.g. fingers on a hand) is insufficient in the accretive growth regime.

Fig.~\ref{fig:t6-1}, \ref{fig:t6-2}, and \ref{fig:t6-3} show the correspondence maps from a source mesh to a target mesh generated by FOLIAGE and baselines from two distinct viewing angles each. As the surface expands and buckles, a small bulge quickly develops into multiple twists and turns which FOLIAGE reliably tracks. Meanwhile, baseline models increasingly lose track of or mismatches features as the morphological complexity of the surface grows under shell physics and material accretion.

\section{SURF-GARDEN Parameters}

\begin{figure*}[ht]
    \centering
    \setlength{\tabcolsep}{2pt}
    \begin{tabular}{cccc}
        \includegraphics[width=0.24\textwidth]{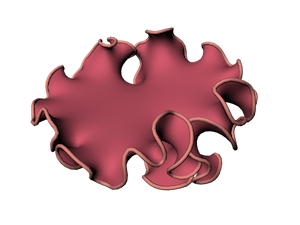} &
        \includegraphics[width=0.24\textwidth]{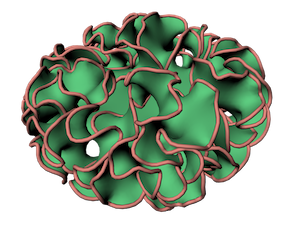} &
        \includegraphics[width=0.24\textwidth]{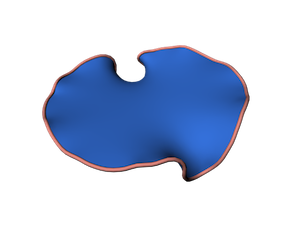} &
        \includegraphics[width=0.24\textwidth]{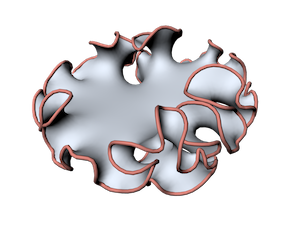} \\
        \includegraphics[width=0.24\textwidth]{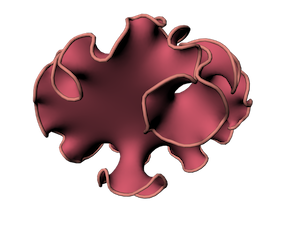} &
        \includegraphics[width=0.24\textwidth]{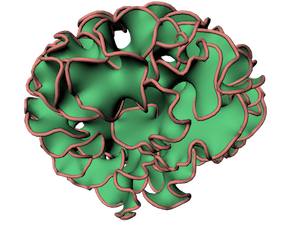} &
        \includegraphics[width=0.24\textwidth]{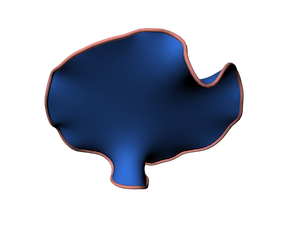} &
        \includegraphics[width=0.24\textwidth]{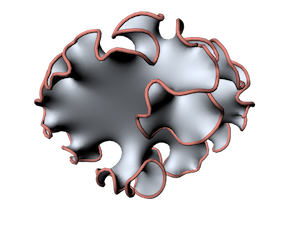} \\
    \end{tabular}
    \caption{Effect of SURF-GARDEN physical control parameters $[k_\text{stretch}, k_\text{shear}, k_\text{bend}]$ (left to right columns): $[0.15, 0.15, 0.25]$, $[0.15, 0.15, 0.2]$, $[0.1, 0.15, 0.2]$, and $[0.15, 0.1, 0.2]$, respectively.}
    \label{fig:dataset-params}
\end{figure*}

SURF-GARDEN supports the exploration of a large morphology space guided by physical control parameters $k_\text{stretch}, k_\text{shear}, k_\text{bend}$. Fig.~\ref{fig:dataset-params} illustrates the effect of their different combinations. A low bending coefficient models a thinner, more flexible surface that is prone to more complex deformations; a higher value models a thicker surface that only permits large-scale deformations to form slowly. Stretching and shearing coefficients further regulate the local behavior of the surface, leading to varied morphology.

\section{Latent Space} 
    
\begin{figure*}[ht]
    \centering
    \includegraphics[width=0.4\textwidth]{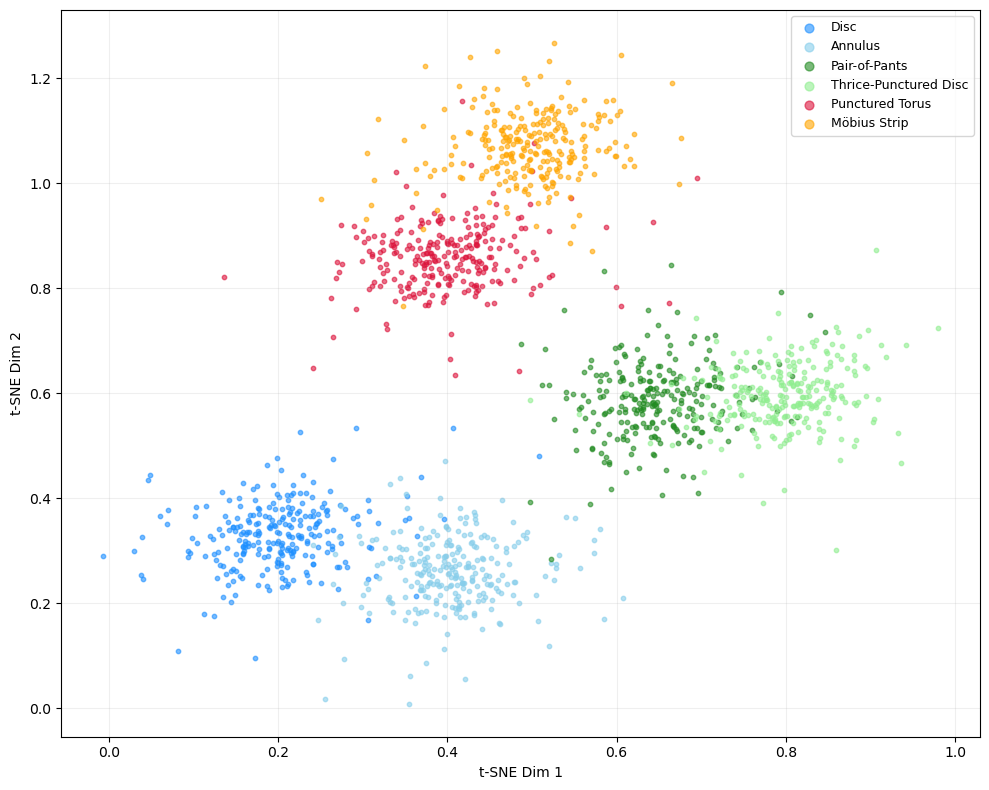}
    \caption{TSNE on topology classes.}
    \label{fig:tsne}
\end{figure*}

\textbf{Topology.} FOLIAGE’s latent space naturally arranges shapes according to their global invariants—genus and boundary count—while not explicitly trained on topology classification tasks. In Fig.~\ref{fig:tsne}, all genus-0 surfaces (disc, annulus, pair-of-pants, thrice-punctured disc) form one region, with boundary-count differences causing small shifts along a shared axis: for example, the disc (one boundary) sits between the annulus (two holes) and the pair-of-pants (three holes). By contrast, the genus-1 punctured torus and the non-orientable Möbius strip form a distinct cluster, reflecting their additional “handle” or “twist.”

\section{Extended Ablation Studies}

Before delving into detailed ablations, we clarify our composite-metric proxy. Rather than tuning four hyperparameters across six individual tasks (and four stress tests), we normalize each task’s evaluation metric into a \([0, 1]\) range (inverting distances so that higher score indicates better performance), weight all tasks equally, and sum them into a single scalar. This composite score strongly correlates with the full multi-task performance of interest (\(\rho \approx 0.92\)), enabling broad five-point sweeps to be run efficiently. Once top-performing settings emerge, we re-evaluate the individual metrics for each task and report them in Tab.~\ref{tab:model-capacity-encoding}, \ref{tab:regularization}, and \ref{tab:optimization}.

\subsection{Statistical Reliability Across Seeds and Topology}
\label{sec:statistical-reliability}

For the experiments most directly supporting hidden-material inference and forecasting, we additionally train five models from independent random initializations. In Tab.~\ref{tab:statistical-reliability}, we report mean $\pm$ standard deviation across trained models; bracketed values are 95\% bootstrap confidence intervals over held-out parent trajectories after averaging predictions across seeds.

The ranking is unchanged across all six SURF-GARDEN topology families. FOLIAGE's C40 ranges from $34.3\pm0.9$ on discs to $52.0\pm1.5$ on Möbius surfaces, whereas uniform allocation ranges from $48.0\pm1.3$ to $82.4\pm2.3$. On the highest-error Möbius family, the respective 95\% intervals are [50.7,53.3] and [80.5,84.4]. Thus, the aggregate forecasting gain is not driven by a single topology family or training seed.

\subsection{Model Capacity and Temporal Encoding}

In Tab.~\ref{tab:model-capacity-encoding}, we swept the latent dimensionality \(d\) from 512 to 1024. Smaller dimensions (512–640) consistently underperform: the model lacks sufficient capacity to encode fine-grained geometric and energetic signals, impairing tasks such as dense correspondence and material regression. Larger dimensions (896–1024) offer diminishing returns—more parameters than data—and exhibit slightly reduced stability during long-horizon rollouts, as the predictor transformer struggles to regularize across a wider channel space. We find \(d = 768\) to be the optimal trade-off, balancing expressivity for physics-informed features (e.g., membrane and flexural energies) with trainability.

We also tuned the EMA (exponential moving average) update rate for the target (privileged-signal) encoder. Slower rates (0.995–0.997) update too sluggishly, causing the context and target embeddings to drift apart, which diminishes the effectiveness of energy-gated message passing. Faster rates (0.999–0.9995) over-smooth the target, preventing it from reflecting the latest context weights, and thereby hamper auxiliary energy regression. We observed that a rate of 0.998 best balances stability and responsiveness.

\begin{table*}[t]
\centering
\begin{tabular}{@{}lclclc@{}}
\toprule
\textbf{Token Drop Ratio} & \textbf{Score} &
\textbf{Modality Drop Ratio} & \textbf{Score} &
\textbf{Action Drop Prob.} & \textbf{Score} \\
\midrule
15\% & 0.78 &
20\% & 0.79 &
0\% & 0.80 \\
20\% & 0.80 &
25\% & 0.81 &
5\% & 0.81 \\
25\% (Ours) & \textbf{0.82} &
30\% (Ours) & \textbf{0.82} &
10\% (Ours) & \textbf{0.82} \\
30\% & 0.80 &
35\% & 0.79 &
15\% & 0.81 \\
35\% & 0.77 &
40\% & 0.75 &
20\% & 0.78 \\
\bottomrule
\end{tabular}
\vspace{0.5em}
\caption{Ablation results for regularization strategies. Each column group shows the effect of sweeping one dropout-related hyperparameter on the composite validation score. The selected configuration for each is highlighted in bold.}
\label{tab:regularization}
\end{table*}

\begin{table}[t]
\centering
\setlength{\tabcolsep}{3pt}
\begin{tabular}{@{}lclclclc@{}}
\toprule
\textbf{Learning Rate} & \textbf{Score} &
\textbf{Weight Decay} & \textbf{Score} &
$\lambda_{\mathrm{energy}}$ & \textbf{Score} &
$\lambda_{\mathrm{vc}}$ & \textbf{Score} \\
\midrule
\(5.0\times10^{-4}\) & 0.78 &
\(5.0\times10^{-3}\) & 0.79 &
0.00 & 0.75 &
0.00 & 0.76 \\
\(7.5\times10^{-4}\) & 0.80 &
\(7.5\times10^{-3}\) & 0.81 &
0.01 & 0.79 &
0.02 & 0.80 \\
\(1.0\times10^{-3}\) (Ours) & \textbf{0.82} &
\(1.0\times10^{-2}\) (Ours) & \textbf{0.82} &
0.02 (Ours) & \textbf{0.82} &
0.04 (Ours) & \textbf{0.82} \\
\(1.5\times10^{-3}\) & 0.79 &
\(1.5\times10^{-2}\) & 0.80 &
0.04 & 0.80 &
0.06 & 0.81 \\
\(2.0\times10^{-3}\) & 0.75 &
\(2.0\times10^{-2}\) & 0.76 &
0.08 & 0.78 &
0.08 & 0.77 \\
\bottomrule
\end{tabular}
\vspace{0.5em}
\caption{Ablation results for optimization and loss weighting. Each group shows a sweep over one hyperparameter and its effect on the composite validation score.}
\label{tab:optimization}
\end{table}

\subsection{Temporal Horizon Sampling}

Choosing the rollout-horizon distribution is critical in a physics-aware world model. As Tab.~\ref{tab:model-capacity-encoding} shows, when we bias sampling toward short horizons (e.g., Uniform(1, 4)), the model learns only incremental dynamics and performs poorly in mid-term predictions; Chamfer and vertex-drift errors spike after 10 steps. Conversely, sampling very long horizons (e.g., Uniform(1, 36)) spreads the model’s capacity across a wide temporal range, weakening both short-term fidelity and long-term coherence. Our Uniform(1, 20) policy emphasizes the early and mid-growth phases—where prediction is most critical—while still exposing the model to challenging, longer-range rollouts. Empirically, we find that this sampling regime consistently maximizes the composite score without overfitting to either extreme.

\subsection{Regularization and Robustness}

\textbf{Token and Modality Dropout.} In our cross-modal fusion setup, we independently drop 25\% of tokens per modality and 30\% of entire modalities. As Tb.~\ref{tab:regularization} shows, lower dropout rates (15–20\% token, 20–25\% modality) fail to regularize adequately: the model overfits to specific sensor patterns and degrades under simulated sensor dropout (Stress S1). Higher rates (30–35\% token, 35–40\% modality) deprive the fusion transformer of coherent correspondence signals, weakening geometry–correspondence alignment and degrading performance on tasks such as topology classification and retrieval. The selected dropout rates strike a balance, simulating realistic sensor failures without removing so much information that cross-modal attention cannot reconstruct object structure.

\textbf{Action Dropout.} We further experimented with dropping the action token during training (0\%–20\%) in Tab.~\ref{tab:regularization}. Omitting action dropout leads to a model that is overly dependent on control inputs and generalizes poorly when such inputs are noisy or absent. Conversely, excessive dropout (15–20\%) enforces robustness at the cost of physical consistency, as the model may ignore legitimate control signals. A moderate 10\% dropout encourages the model to infer actions from observed state transitions, while still forming tight action–perception loops when control signals are present.

\section{Optimization}

\textbf{Learning Rate and Weight Decay.} As Tab.~\ref{tab:optimization} shows, a low AdamW learning rate (e.g., \(5 \times 10^{-4}\)) leads to slow convergence and under-optimized parameters, while a high learning rate (e.g., \(2 \times 10^{-3}\)) causes unstable gradients, especially in the multi-head self-attention layers of the predictor. Similarly, a weak weight decay (e.g., \(5 \times 10^{-3}\)) under-regularizes the high-dimensional latent space, whereas overly strong decay (e.g., \(2 \times 10^{-2}\)) suppresses meaningful emergent physics representations. Our chosen configuration—learning rate of \(1 \times 10^{-3}\) and weight decay of \(1 \times 10^{-2}\)—yields smooth optimization and robust generalization.

\textbf{Energy and Variance–Covariance Loss Weights.} The auxiliary energy regression weight $\lambda_{\mathrm{energy}}$ and variance–covariance regularizer $\lambda_{\mathrm{vc}}$ govern how much the model prioritizes privileged physical signals over raw rollout accuracy. Setting the energy weight to zero $\lambda_\mathcal{E} = 0$ causes material inference to degrade, while excessive weight (e.g., \(\lambda_{\mathrm{energy}} = 0.08\)) pulls the latent space toward physics features at the cost of open-loop prediction accuracy, worsening Chamfer and drift metrics. We select \(\lambda_{\mathrm{energy}} = 0.02\) and \(\lambda_{\mathrm{vc}} = 0.04\) to ensure that physics cues meaningfully inform the representation without overwhelming the primary learning signal, striking a balance between interpretability and predictive performance.

\clearpage


\end{document}